\documentclass{article}

\usepackage{arxiv}

\usepackage[utf8]{inputenc} 
\usepackage[T1]{fontenc}    
\usepackage{hyperref}       
\usepackage{url}            
\usepackage{booktabs}       
\usepackage{amsfonts}       
\usepackage{nicefrac}       
\usepackage{microtype}      
\usepackage{lipsum}

\usepackage{graphicx}
\usepackage{amssymb}
\usepackage{amsmath}
\usepackage{graphicx}
\usepackage{csquotes}
\usepackage{mwe}
\usepackage{subfig}
\usepackage{listings}
\usepackage{multirow}
\usepackage{url}
\usepackage{hyperref}
\usepackage[ruled,vlined]{algorithm2e}

\usepackage{color}

\definecolor{dkgreen}{rgb}{0,0.6,0}
\definecolor{gray}{rgb}{0.5,0.5,0.5}
\definecolor{mauve}{rgb}{0.58,0,0.82}

\title{LioNets: A Neural-Specific Local Interpretation Technique Exploiting Penultimate Layer Information}

\author{
  Ioannis Mollas\\
  Aristotle University of \\Thessaloniki, 54636, Greece\\
    \texttt{iamollas@csd.auth.gr}\\
     \And
       Nick Bassiliades\\
  Aristotle University of \\Thessaloniki, 54636, Greece\\
             \texttt{nbassili@csd.auth.gr}\\
             \And
               Grigorios Tsoumakas\\
  Aristotle University of \\Thessaloniki, 54636, Greece\\
             \texttt{greg@csd.auth.gr}\\

}

\begin{document}
\maketitle

\begin{abstract}
Artificial Intelligence (AI) has a tremendous impact on the unexpected growth of technology in almost every aspect. AI-powered systems are monitoring and deciding about sensitive economic and societal issues. The future is towards automation, and it must not be prevented. However, this is a conflicting viewpoint for a lot of people, due to the fear of uncontrollable AI systems. This concern could be reasonable if it was originating from considerations associated with social issues, like gender-biased, or obscure decision-making systems. Explainable AI (XAI) is recently treated as a huge step towards reliable systems, enhancing the trust of people to AI. Interpretable machine learning (IML), a subfield of XAI, is also an urgent topic of research. This paper presents a small but significant contribution to the IML community, focusing on a local-based, neural-specific interpretation process applied to textual and time-series data. The proposed methodology introduces new approaches to the presentation of feature importance based interpretations, as well as the production of counterfactual words on textual datasets. Eventually, an improved evaluation metric is introduced for the assessment of interpretation techniques, which supports an extensive set of qualitative and quantitative experiments.
\end{abstract}

\keywords{Interpretable \and Explainable \and Machine Learning \and Neural Networks}

\section{Introduction}

Interpretable machine learning (IML) aims to discover the rationale behind the decision of a system, and explainable machine learning aims to present this kind of analysis in a human-understandable and convincing manner. This is one of the many definitions of IML presented in the literature~\cite{du2019techniques, Doshi-velez2017, rudin2019stop, gilpin2018explaining}. IML is currently in the spotlight of artificial intelligence research. It is a promising field that aims to address several key socio-economic and ethical issues that machine learning systems may create~\cite{du2019techniques}. For example, interpretable machine learning systems can assist underwriters in the insurance and banking sectors~\cite{bussman2019explainable}, and provide explanations as to why someone's right to free speech has been trampled by an automated process of a social network~\cite{wang2018interpreting}. In addition, interpretable machine learning is the key to transform efficient machine learning procedures, such as predictive maintenance~\cite{Giurgiu2019}, into more descriptive and reliable ones, like prescriptive maintenance~\cite{khoshafian2015digital}.

Currently, often, we are confronted with phenomena of opaque machine learning systems that do not give the executives of businesses the confidence to authorise their deployment within their organisations, despite being able to solve a lot of problems more efficiently and precisely than humans. That's why a lot of companies are willing to invest in research in the field of IML, empowering academic efforts. For instance, a number of companies produced deep-learning oriented facial recognition products, which were described as unreliable, during an investigation performed by the National Institute of Standards and Technology\footnote{\url{https://bit.ly/31Q5kLV}} revealing racial and gender biases, should take immediate measures to prevent these phenomena from occurring in their future products. Another important factor enabling businesses to invest in research methodologies for interpretable systems is the need to comply with legislation such as the EU General Data Protection Regulation (GDPR)~\cite{gdpr} and the US Equal Credit Opportunities Act\footnote{ECOA 15 U.S. Code §1691 et seq.: \url{https://www.law.cornell.edu/uscode/text/15/1691}}.

Most interpretable machine learning techniques focus on the concept of local explanation~\cite{Ribeiro2016quotWhyClassifier}. In contrast to the global explanation that deals with the structure and logic of a model, the local explanation concerns the prediction of a single instance. In addition, model-agnostic and model-specific methodologies can be identified where the former can be applied indifferently to any machine learning model, while the latter concerns methodologies for a particular family of machine learning models or even a specific architecture.

An intuitive technique to agnostically derive interpretations from an obscure model is to train a transparent model, such as a decision tree or a linear model, using the original data and the predictions of the obscure model, constructing a surrogate model. That would be a global interpretation attempt. Nevertheless, given the low capacity of transparent models, as opposed to obscure models that are well known for their high capacity, this interpretation attempt would fail to replicate the true logic of the complex model. On the other hand, by creating custom sub-spaces (neighbourhoods) of training data around a given instance, we might be able to fit more efficiently a transparent model, creating a local surrogate model to discover the features of that instance that influenced its prediction. Few model-agnostic techniques, which adopt the concept of the local surrogate model, in order to adapt to different data types, they need to modify both the generation process and the interpretation process. Indeed, when the data are textual or other sparse type data, these methods face some problems with their neighbourhood generation processes~\cite{mollas2019lionets}.

Even if the use of model-agnostic approaches seems practical and preferable over model-specific approaches, in those cases where researchers have the ability to use the inner structure of the model they are explaining, all this knowledge remains unexploited. Thus, model-specific approaches can leverage this information to provide better and more reliable explanations for particular models, such as neural networks that advance in tasks of object detection~\cite{neuralObject} or machine translation~\cite{neuralTranslation}, among others~\cite{neuralGeneral}.

Techniques for local explanation of neural networks attempt to provide useful information about the influence of the input to the output. A family of techniques, like Gradient$\times$Input~\cite{gradXinput}, Guided Backpropagation~\cite{guidedBack}, Grad-Cam~\cite{gradCam}, Layerwise Relevance Propagation (LRP)~\cite{LRP}, Deep Taylor Decomposition (DTD)~\cite{montavon2017explaining} and DeepLift~\cite{deepLift}, propagate the influence of a signal backwards through the layers from the output neuron to the input, in one pass. However, most of these methods make assumptions concerning the type of activation functions and the network's architecture~\cite{ancona2018towards}. Guided Backpropagation is limited to Rectified linear units (ReLU)~\cite{relus}, LRP to activation functions where $f(0) = 0$ and Grad-Cam to convolutional neural networks.

In this work, we present a local neural-specific interpretation technique, firstly introduced in our previous work~\cite{mollas2019lionets}, known as LioNets (Local Interpretation Of Neural nETworkS through penultimate layer decoding). LioNets aims to build a local transparent model to interpret an instance's prediction. To train this model, it constructs a local neighbourhood at the penultimate layer of the neural network, leveraging the rich semantic information that this layer contains. The generation process is the same regardless of the shape of the input, but we need to have neighbour representations in the original input space to train the transparent model. To achieve this, LioNets requires a decoder that is capable of reconstructing examples from their abstract representations to original space representations. 

This paper expands the local interpretation technique of LioNets, by contributing the following:
\begin{itemize}
    \item A demonstration of LioNets implementation on a variety of data types
    \item A collection of new approaches to present interpretations of time series data
    \item A new method for providing counterfactual words on textual datasets
    \item An extended version of the original deterministic process for generating neighbours for an instance
    \item A qualitative and quantitative evaluation, through a comparison between different explanation techniques
    \item An extensive presentation of LioNets theory using a toy example
\end{itemize}

\section{Related Work}
Interpretable machine learning refers to the ability of machine learning models to provide users with useful insights about their structure and decisions. A machine learning model can be inherently interpretable, like linear models~\cite{LRigre}, generalised linear models~\cite{McCullagh1972GeneralizedLM}, decision trees~\cite{DTrees}, bayesian models~\cite{NB1, NB2} or even $k$-nearest neighbours models~\cite{knnAltManx}, or uninterpretable like random forests~\cite{RForests}, support vector machines~\cite{SVM} or neural networks~\cite{FNN}.

Techniques that seek to shed light on the rationale of a machine learning model (uninterpretable or not) can be classified based on their adaptability as model-agnostic and model-specific approaches. Model-agnostic techniques aim to interpret any model of machine learning indifferently, while model-specific techniques focus on a particular family of machine learning algorithms. Furthermore, the interpretation techniques can be distinguished in global and local-scopic approaches. However, given that there are so many techniques available, it is more important to measure and select the most effective one according to certain metrics. 

\subsection{Model-Agnostic Approaches}
Model-agnostic techniques are attempting to provide interpretations of any machine learning model indifferently. Some methodologies, such as feature importance~\cite{permFI}, partial dependence plots (PDP)~\cite{pdp} and individual conditional expectation (ICE) plots~\cite{iceplots}, rely on input data permutation methods and attempt to present the effect of each feature in a global-scope. SHAP~\cite{shap}, a technique that takes advantage of shapley value computation, provides both global and local explanations for any machine learning model. On the other hand, techniques like LIME~\cite{Ribeiro2016quotWhyClassifier} and Anchors~\cite{RibeiroAnchors:Explanations}, they use instance-level permutations, which along with the predictions of the machine learning model trying to interpret, are given as input to an interpretable model. Such a model is called a surrogate model, and is usually a decision tree or a linear model.

LIME~\cite{Ribeiro2016quotWhyClassifier} is a state-of-the-art methodology of explaining machine learning predictions, as already described. For one particular textual instance, LIME generates a neighbourhood of a specific size by arbitrarily choosing to set a zero value in one or more features. Then, the cosine similarity of each neighbour to the original instance is calculated and multiplied by one thousand. This will be the weights on which the simple linear model will depend during its learning phase. Consequently, the most similar neighbours will have a greater impact on the training process of the linear model. In the case of sparse data, we can detect a drawback of LIME. Because of the perturbation method occurring on the original space, LIME can only produce $2^n$ unique neighbours, where $n$ is the number of non-zero elements. For instance, in textual data, the non-zero features are just six in a sentence of six words expressed as a vector of four thousand features, each of which corresponds to one word in the vocabulary. Thus, it can produce just $2 ^ 6 = 64$ separate neighbours. Nevertheless, LIME can create a neighbourhood of five thousand instances by random sampling from 64 unique neighbours.

When handling tabular data, LIME follows a different generation process. This procedure depends heavily on the training set to identify the distribution of features by extracting statistics for each of them. The permutations around an instance are then configured with respect to the centres of the distributions of each feature. Similarity computation is even more important in this phase of generation of the neighbourhood, as it is the only element assessing the locality among the synthetic instances and the original instance.

Another model-agnostic local-based approach is called Anchors~\cite{RibeiroAnchors:Explanations}. The Anchors approach introduces several improvements over LIME. The form of the interpretation of anchors is textual and has the shape of a rule. For each instance, a single human-readable rule is created and presented to the user, followed by precision and local coverage scores. Nonetheless, the approach is subject to the need for highly engineered setups, where the perturbation mechanism should be specifically designed for each scenario. In fact, hyper-parameter tuning is required to obtain concrete explanations while, at the same time, there is a lack of evaluation of how meaningful the explanations are, making the whole process very difficult.

Shapley's~\cite{shapley1,shapley2} values are a game theory inspired technique that defines how much each ``player’’ has contributed to the outcome of a collaborative game. In machine learning, the ``player’’ is a feature’s value, while the collaborative game is the decision-making process. SHAP \cite{shap} combines LIME’s idea about sub/local spaces to compute the Shapley values for an instance. This kind of processed information is presented in a feature-importance fashion to the end-user. In addition, SHAP can also offer global explanations through a diversity of plots. At the same time, SHAP has a lot of variations, which focus on particular problems, like TreeExplainer, GradientExplainer, DeepExplainer and KernelExplainer. A negative aspect of SHAP is its heavy computational nature.

X-SPELLS~\cite{Orestis} is a methodology concerning neighbourhood generation for instances of a textual corpus, in order to create meaningful neighbours. Then, interpretation techniques, such as surrogate decision trees, will use the created neighbours to extract explanations. However, this technique has been proposed as a data-specific approach. The neighbourhood generation process is based on a variational autoencoder (VAE) that is trained on a particular dataset without being influenced by the structure of the machine learning model rather than its output. Moreover, utilising decision trees as surrogate models, they extract sets of counterfactual words for an instance's prediction.

\subsection{Neural-Specific Approaches}
A set of neural-specific approaches are the backpropagation-based techniques, which measure the importance for all input features in a single backward pass through the network. Although, these methods are generally faster than perturbation-based methods, due to nonlinear saturation, discontinuous and negative gradients, their results may be inaccurate.

In saliency maps~\cite{saliencyMaps}, the gradient of the output probability of a network with respect to the input is calculated through back-propagation, producing a ``heatmap'' or a ``saliency map''. Gradient$\times$Input~\cite{gradXinput} technique multiplies the gradient computed by the saliency maps, with the original input, addressing the ``gradient saturation'' problem, possibly appearing in the ``heatmaps''. A disadvantage of Gradient$\times$Input method concerns the probability of ``noisy'' explanations due to the ``shattered'' gradients of a deep neural network~\cite{samekexplainable}.

Layer-wise Relevance Propagation (LRP)~\cite{LRP} is a model-specific local-based technique which identifies essential features, running a backward pass on a neural network. In this sense, the LRP technique redistributes the output, exploiting the gradients, backwards over the network to calculate the nodes' contribution to the instance's prediction. LRP is an explanation technique, which rests upon the theoretical foundations of the DTD~\cite{montavon2017explaining}. When all neural network activation functions are ReLUs, then LRP is equivalent to Gradient$\times$Input. Nevertheless, when LRP is applied to architectures that contain Sigmoid, Softplus~\cite{softplus}, or other nonlinear activation functions, where $f(0)\neq 0$ may generate numerically unstable interpretations~\cite{ancona2018towards}.

DeepLift~\cite{deepLift} is another technique relying on backpropagation assigning significant scores to the features of the input. The approach is based on measuring the difference between the output of the example (to be explained) and the output of a specified `reference’, as well as the difference between the input of the example and the input `reference'. DeepLift addresses the problem of model saturation, and at the same time, it does not assign misleading importance scores to biases. A drawback of DeepLift is that identifying the most suitable references demands domain knowledge. Additionally, when applied to recurrent neural networks, it fails to produce meaningful results~\cite{ancona2018towards}. An interesting fact is that SHAP incorporates DeepLift algorithm to calculate the SHAP values for the deep learning models approximately.

Finally, the attention-based approaches are another set of neural-specific interpretation methodologies. In the field of natural language processing (NLP), a ground-breaking approach, the attention mechanism~\cite{attention1}, has been introduced to address a variety of performance issues. Attention mechanism introduces a context layer to the neural network that tends to assign an indication for the relationship between the input and the target. A number of works have converted the attention layer information exploited into interpretations~\cite{attention2}, and even visualised this knowledge~\cite{attention3}, having created a different interpretation family. However, studies have shown that such approaches can provide noisy factors of importance, identifying these types of techniques as unreliable methods~\cite{antiAttention}.

\subsection{Evaluation of Interpretation Techniques}
\label{sec:withFaith}
Interpretable solutions based on feature importance have been on the spotlight for a while. There are a few metrics, such as \textit{fidelity} or \textit{number of non-zero weights}, which are currently the most common choices for researchers. In practice, however, these metrics cannot present the superiority of an algorithm against competitive techniques. Metrics such as \textit{robustness} and \textit{fairness} have therefore emerged. A meta-explanation tool based on argumentation, and influenced by fairness, able to be utilised as an evaluation metric was introduced as \textit{Altruist}.

\emph{Fidelity} measures the ability of a transparent model to `imitate' the decisions of an obscure regression or classification model. Fidelity is measured in both global and local-scopic aspects. For a dataset $D=[(x_1,y_1),\dots,(x_n,y_n)]$ and a trained machine learning model $f(x)$ we want to explain, we measure the fidelity of a transparent model $g(x)$ in a subset $D'\subseteq D$ as follows ($D'$ can be a local neighbourhood): 

\begin{equation}
    \label{eq:fidelity}
    fidelity(f,g,D') = 1 - \frac{1}{|D'|}\sum_{i}^{|D'|}|g(x_i) - f(x_i)|
\end{equation}

Another useful metric many researchers use in their evaluation experiments is the \emph{average number of non-zero weights} of the explanations provided by their systems. For a transparent system providing explanations like $e(x_i) = \{f_j = influence | f_j \in F, influence\in \mathbb{R}, influence \neq 0\}$, where $F$ is the feature set of the dataset, the average number of non-zero weights is given from the expressions in Eq.~\ref{eq:anzw}. We can use this metric to measure both local and global explanations. 
The best model in terms of this metric, is the one with the smallest average. 

\begin{equation}
    \label{eq:anzw}
    average\_non\-zero\_weights = \frac{1}{|D'|}\sum_{i}^{|D'|}|e(x_i)| \text{\\}
\end{equation}{}

Based on Lipschitz's continuity, robustness~\cite{robustness} investigates how different the explanations given for two examples of subtle divergence are. This way, the instability of an interpretation technique could be uncovered. More specifically, robustness relies on the neighbourhood-based local Lipschitz continuity. Robustness is seen in Eq.~\ref{eq:robustness}, attempting to find the divergence for an explanation of a particular instance $x_i$, and an explanation of a neighbour of $x_i$, $x_j \in B_{\epsilon}(x_i)$, maximising the difference between the two explanations ($f_
{expl}$) and the concepts assigned to each one of these two instances ($h$). The concept space is either provided by the model designer or learned through the training process. However, it is more difficult to define this concept space for black boxes that have already been trained. Essentially, this metric calculates the instability of the explanation technique.

\begin{equation}
    \label{eq:robustness}
    robustness(x_i)=\arg\max_{x_j\in B_{\epsilon}(x_i)} \frac{||f_{expl}(x_i)-f_{expl}(x_j)||_2}{||h(x_i)-h(x_j)||_2}
\end{equation}

The faithfulness metric was introduced in an experimental setup to evaluate the different explanation techniques applied in neural networks containing recurrent network layers on a binary classification task in a textual dataset~\cite{faithfulness}. As it is visible in Eq.~\ref{eq:faith}, where $L$ is the total number of instances, this metric compares the prediction probability between the original instance before ($Probability(x_{i}^{original})$) and after ($Probability(x_{i}^{tweaked})$) removing the most important feature by setting its value to zero. In this particular study, the features' importance were determined for each sentence of a paragraph and the most important sentence was omitted from the paragraph in order to determine the faithfulness of the interpretation. The explanation approach with the highest degree of faithfulness is considered to be the best technique.

\begin{equation}
    \label{eq:faith}
    faithfulness = \frac{1}{L}\sum_{i=1}^{L}(Probability(x_{i}^{original}) - Probability(x_{i}^{tweaked})) \text{\\}
\end{equation}

Finally, Altruist~\cite{altruist}, heavily influenced by the faithfulness metric, introduces the concept of truthfulness. An importance $z_j$ assigned to a feature $f_j$ is considered truthful when the expected changes to the output of the predictive model are correctly observed with respect to the changes that occur in the value of this feature. This metric is applied on interpretation techniques providing feature importance $Z$, judging each feature importance $z_j \in Z$ as truthful ($z_j^t$) or untruthful ($z_j^u$). As shown in Eq.~\ref{eq:alt}, where $L$ is the total number of instances, the mean average number of untruthful feature importances per instance is the final score.

\begin{equation}
    \label{eq:alt}
    altruist =  \frac{1}{L}\sum_{i=1}^{L}(|[z_j^u|z_j^u \in Z', j\in[0,Z']]| ) \text{\\}
\end{equation}

\section{LioNets}
\begin{figure}[ht]
\centering
\includegraphics[width=\textwidth]{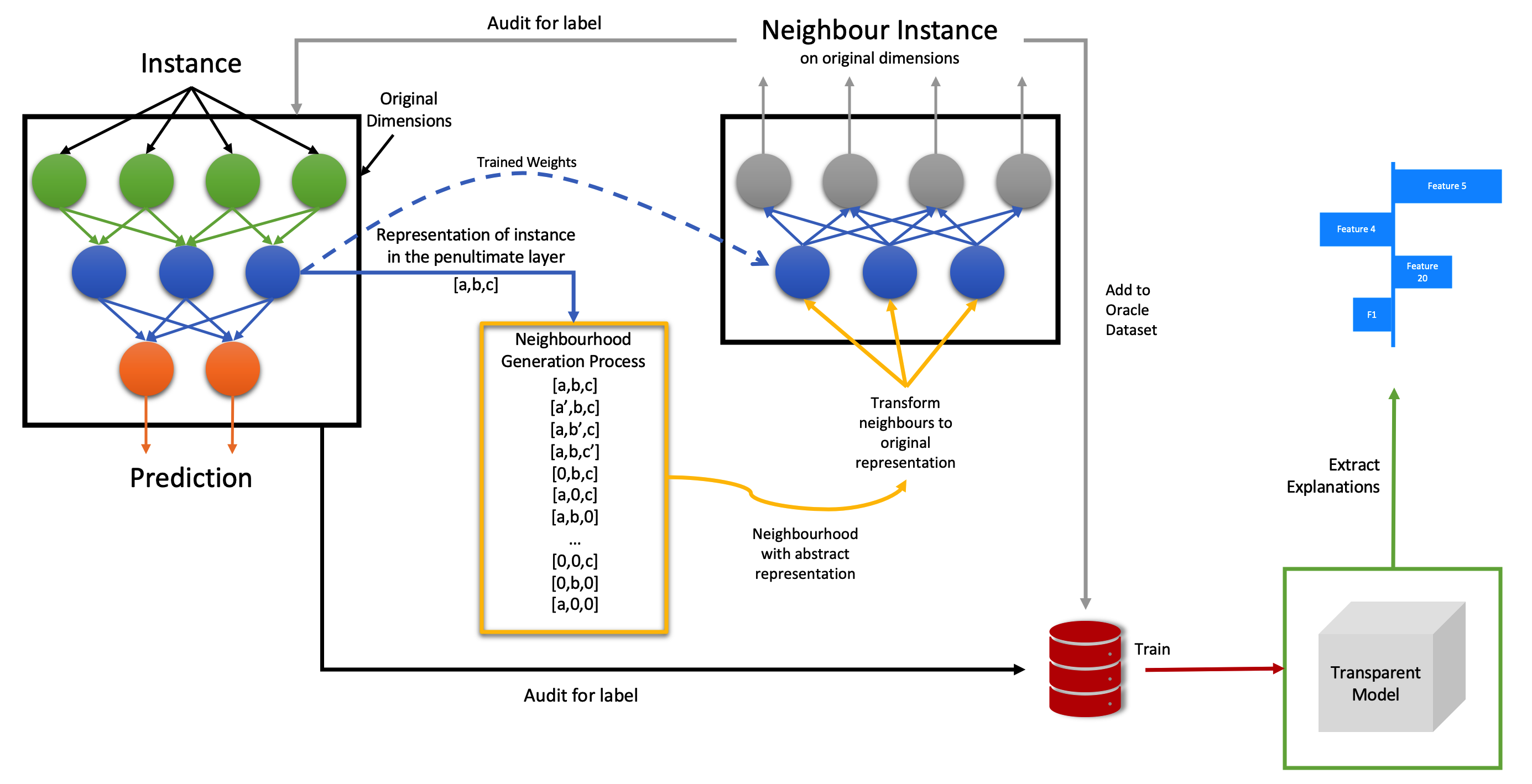}
\caption{LioNets' architecture. There are four fundamental mechanisms: a) the predictor, b) the decoder, c) the neighbourhood generation process and d) the transparent model} \label{lioNetsArchitecture}
\end{figure}

LioNets is a local-based model-specific interpretation technique for neural network predictors (NN). LioNets take advantage of NNs by exploiting the latent information via the penultimate layer (encoded representation) and the output for a given instance (predictions). Specifically, through the multi-informative penultimate layer of a neural network called abstract or latent space, LioNets creates for an instance neighbours who are semantically closer to the instance to explain its prediction, by training a local transparent linear model. The neighbours would however have abstract representation. In order to get predictions for these neighbours, they must be converted into the original space by a decoder. Then, via the NN's predictions for the transformed neighbours, a local transparent model will be trained and the interpretation will be extracted. This process is presented in the architecture shown in Figure~\ref{lioNetsArchitecture}.

\subsection{Neighbourhood Generation Process}
Acquiring the encoded representation of a given instance by the NN, the second component of LioNets is the neighbourhood generation process (NG). This component will create a neighbourhood around the encoded instance. In the first implementation of LioNets, the neighbourhood generation process was a deterministic process, which was creating a strange neighbourhood distribution in both latent and original space. In this work, we are taking advantage of an extended version of the original deterministic process. Nevertheless, it is easy to apply any other known neighbourhood generation process.

\begin{algorithm}[ht]
 \KwIn{encoded\_instance, number\_of\_neighbours, features\_stats}
 \KwOut{new\_value}
   dimensions\_ $\gets$ dimensions\_of(encoded\_instance) \\
   neighbours $\gets [ ]$ \\
   \For{$i \in [0, dimensions\_]$}{
       instance\_copy $\gets$ copy\_of(encoded\_instance)\\
       value $\gets$  instance\_copy[i] \\
       \For{$level \in [normal, weak, strong]$}{
            instance\_copy[i] $\gets$ determine\_value(value, features\_stats[i], level)\\
            neighbours.add(instance\_copy) 
    }
   }
   \While{size\_of(neighbours) $\leq$ number\_of\_neighbours} {
    instance\_copy $\gets$ copy\_of(encoded\_instance)\\
    \For{i $\in$ random\_binary\_vector(dimensions\_).nonzero()}{
     instance\_copy[i] $\gets$ determine\_value(value, features\_stats[i], weak)\\
    }
    neighbours.add(instance\_copy)
   }
   neighbours $\gets$ neighbours[:number\_of\_neighbours]\\
   \KwRet{neighbours}
 \caption{Neighbourhood generation process}
 \label{alg:neighbours}
\end{algorithm}

The size $N$ of the neighbourhood is determined by the user manually. The preferred size is at least $2000$ neighbours or more when the abstract space size is over $500$ dimensions, as emerged by our experiments. Depending on the abstract space size $L$, the generation process will create first order neighbours, which will be in population three times the dimensions of the abstract space, and differ by only one feature from the abstract representation of the original instance. The Algorithm~\ref{alg:neighbours} describes the generation process. The new value for each instance is determined with respect to the abstract feature's distribution across all training instances, as shown in Algorithm~\ref{alg:new_f_v}, and it will be generated sampling Gaussian noise.

\begin{algorithm}[ht]
 \KwIn{value, i\_features\_stats, level}
 \KwOut{new\_value}
    i\_min $\gets$ i\_features\_stats[0],
    i\_max $\gets$ i\_features\_stats[1] \\
    i\_mean $\gets$ i\_features\_stats[2],
    i\_std $\gets$ i\_features\_stats[3] \\
   \uIf{level is weak}{
     $i\_std \gets \frac{i\_std}{2}$\\
  }
  \ElseIf{level is strong}{
     $i\_std \gets i\_std \times 2$\\
  }
   noise $\gets$ gaussian\_noise(i\_mean, i\_std)\\
   \KwRet{min(max(value+noise,i\_min),i\_max)} 
   
 \caption{Process of determining new value for a feature}
 \label{alg:new_f_v}
\end{algorithm}

For each abstract feature, three different Gaussian noises will be produced. The first Gaussian noise will be generated with the standard deviation of the feature's distribution, the second with the half value of the standard deviation in order to limit the noise (weak noise), and the third with the double value of standard deviation to generate stronger noise (strong noise). The noises will be added to the existing value of the feature, and if the new values are greater than the maximum value or less than the minimum value, they will be changed to the maximum and the minimum value, respectively, so that the new neighbours remain within the range of the feature's distribution. Thus, for an abstract representation of $L$ dimensions, we acquire $3 \times L$ neighbours. 

If $3\times L$ neighbours are not sufficient for the desired neighbourhood size $N$ we proceed to the second-order neighbours. Till we collect $N - 3\times L$ neighbours, we generate a randomised binary vector of $L$ dimensions, and for the non-zero dimensions, we create a weak noise (as mentioned in Algorithm~\ref{alg:new_f_v}) for these dimensions, and we add it to the original instance's abstract representation.


The neighbourhood generation process has a significant impact on the interpretation of the prediction. The idea behind local explanations is to discover representative sub-spaces, a.k.a. neighbourhoods. Then, in these smaller spaces, non-linear relations may be absent, thus transparent models will be able to capture the neighbourhood's most important features, concerning the black-box model's decisions. To achieve this, the NN component will assign target values (predictions) to each neighbour in the form of probabilities, in classification tasks, or real continuous values in regression tasks. Although in order to do this, we need to have the representation of each of the neighbours in the original space. This is going to happen through decoding, as explained in the following section.

\subsection{Neighbourhood Decoding Process}
After generating neighbours through the NG component, we will need to decode them to the original space through a decoder to train the local transparent model. Therefore, the following component of LioNets is the neighbourhood decoding process (ND). A way to acquire the representation of the neighbours in the original input space is by using decoders. However, training decoders is a much more complex task than training predictive models. The task of a decoder is to inverse transform an instance from one representation to its original representation. Due to the fact that the multiplications inside a neural network through encoding most probably will contain non-square matrices, thus non-invertible matrices, the task of a decoder will be to create pseudo-invertible matrices to transform the data. This is a difficult task, as well as computationally-heavy. Moreover, it is directly affected by the data and the trained predictor/encoder.

By training a classifier and extracting the encoder, the decoder maps $h$, the encoded representation of an instance, to the reconstruction $x'$, the original representation of the instance, of the same shape as $x'=\sigma'(W'h+b')$, where $\sigma'$, $W'$ and $b'$ for the decoder may be unrelated to the corresponding $\sigma$,$W$, and $b$ for the encoder. These models are trained to minimise reconstruction errors, often referred to as ``loss'' function, like the mean absolute error (Eq.~\ref{eq:mae}) or the mean square error (Eq.~\ref{eq:mse}). Another interesting loss function that is used to train decoders and autoencoders is the binary cross-entropy (Eq.~\ref{eq:bc})~\cite{creswell2017denoising} or the Kullback-Leibler divergence~\cite{kingma2013auto}.

\begin{equation}
\label{eq:mae}
    {\mathcal {L}}_{mae}(\mathbf {x} ,\mathbf {x'} )=\|\mathbf {x} -\mathbf {x'} \|
\end{equation}{} 

\begin{equation}
\label{eq:mse}
    {\mathcal {L}}_{mse}(\mathbf {x} ,\mathbf {x'} )=\|\mathbf {x} -\mathbf {x'} \|^2
\end{equation}{} 

\begin{equation}
\label{eq:bc}
    {\mathcal {L}}_{binary\_cross\text{-}entropy}(\mathbf {x} ,\mathbf {x'} )= -\mathbf{x} log(\mathbf{x'}) - (1-\mathbf{x})log(1-\mathbf{x'})
\end{equation}{} 

As we have mentioned, training a decoder is a challenging task, and each problem needs a different architecture for the decoder. A general way to design a decoder is to use the predictor's inverse architecture approximately. Examples of building decoders for different problems can be found in the section~\ref{sec:implementation}. From our experiments, we concluded that it is easy to build a decoder on textual data, when using TFIDF representations, or time-series data, using an architecture similar to the inverted predictor's architecture. However, it was more difficult to train the decoder successfully in text classification with embeddings, and a lot of experimentation was required.

\subsection{Why Latent-Space Neighbourhood Encoding and Decoding?}
Before proceeding to the next and final component, we argue, by way of an example, why the generation of neighbours in latent space is better. Given a dataset $D$ containing $N$ instances $x_i$, $D = \{x_i,y_i|i\in [0,N-1]\}$, each instance has a specific number of $M$ features $x_i = [f_{i,0}, f_{i,1}, \dots, f_{i,M-1}]$. Observing Figure~\ref{lioNetsArchitecture}, a simple neural network architecture, we can see that an instance $x_i \in \mathbb{R}^m$ given as input, in the penultimate layer is $x^{enc}_i \in \mathbb{R}^k$, where $k$ is the number of the nodes, namely the dimension of the penultimate layer. 

The task of data generation is a controversial topic, and by design neighbourhood generation around an instance, which is widely used in explainable machine learning, is not defined properly. Indeed, we can ask: \emph{How do we define neighbourhoods in data?} or \emph{How do we measure adjacency?} A common metric to measure adjacency is the cosine similarity or the euclidean distance~\cite{HAN201239}. However, an open problem in this topic is the curse of dimensionality~\cite{indyk2004nearest}, which raises awareness on whether we are able to trust adjacency in high dimensions or not. Following the advice of researchers working on this problem, it is better to create neighbours in low-dimension spaces. The size of the latent space of a neural network, its penultimate layer representation, is most of the time smaller than the size of the input space. This may help to address the problem of the curse of dimensionality.


\begin{figure}[ht]
\centering
\includegraphics[width=0.74\textwidth]{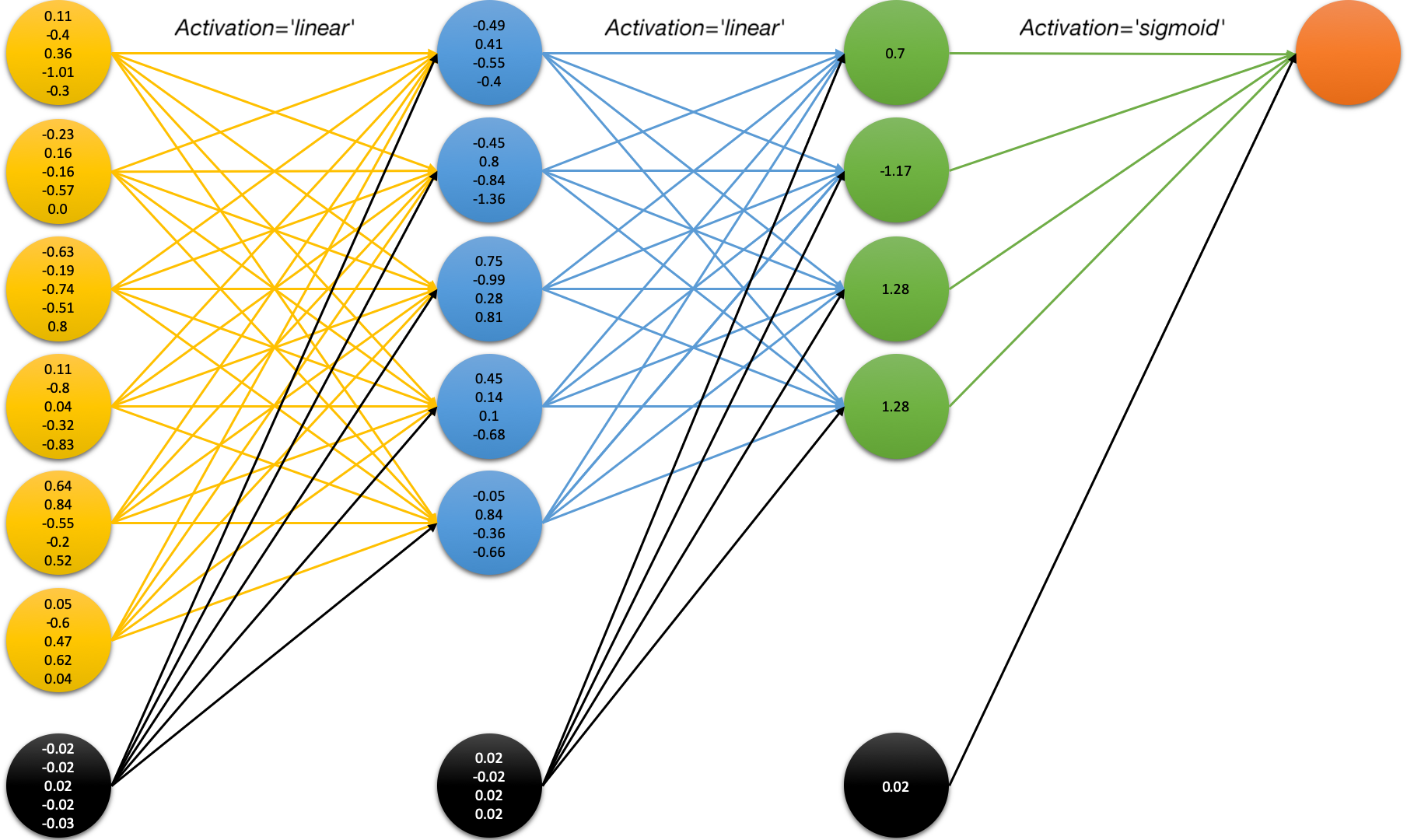}  
\caption{A simple neural network architecture for binary classification} \label{simpleNet}
\end{figure}

In this example (Figure~\ref{simpleNet}) we have 6+1 Dimensions on Input Layer (6 Features per instance and 1 Bias) and 4+1 Dimensions on the Penultimate Layer (Higher-Dimensional Input Space than Latent Space). This neural network has been trained on a binary classification toy problem generated using the \emph{make\_classification} function of Scikit-Learn~\cite{sklearn}, with 100 samples, 6 features and 2 classes.

For an instance $x_i \in \mathbb{R}^6$ we create a neighbour $g_i \in \mathbb{R}^6$, and during the classification process we can take the representation of both examples in the penultimate layer, which are going to be $x^{enc}_i \in \mathbb{R}^4$, and $g^{enc}_i \in \mathbb{R}^4$. There is a probability $x^{enc}_i$ to be equal with $g^{enc}_i$, while $x_i\equiv g_i$, because of the high internal complexity of the neural network, which may eliminate a feature's importance. We want such examples, but in order to discover local neighbourhoods it would be more meaningful to create a neighbour of $x^{enc}_i$ directly. 

Let's have an example inspired by the network in Figure~\ref{simpleNet}. Creating a neighbour in the original space we are going to explore the latent space representation of it. For the instance $x_i = [.18, .0, .63, .24, .58, .81]$, we generate neighbours using LIME. Using for example a neighbour $g_i = [.11, .35, .58, .24, .6, .94]$, with $cosine\_distance=.04$ and $euclidean$ $\_distance$ $= .38$, we observe that in the encoded space the representations of the two instances (the original and the neighbour) are almost identical $x^{enc}_i = [-.17, .41, .03, .08]$, while $g^{enc}_i = [-.17, .41, .02, .08]$. Of course, we cannot say that such an example is not useful, but we definitely can say that such phenomena can ruin the neighbourhood distribution, as it is described in the following paragraph. Thus, it is preferable to create the neighbourhood in the encoded space to ensure both better adjacency and distribution.

\begin{figure}[ht]
\caption*{Distribution of Euclidean distances of neighbours:}
\centering
\minipage{0.43\textwidth}
  \includegraphics[width=\linewidth]{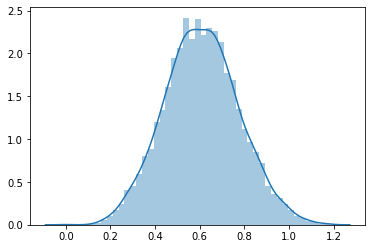}
  \caption{generated on original space}\label{fig:eu_lime}
\endminipage \hspace{0.06\textwidth}
\minipage{0.43\textwidth}
  \includegraphics[width=\linewidth]{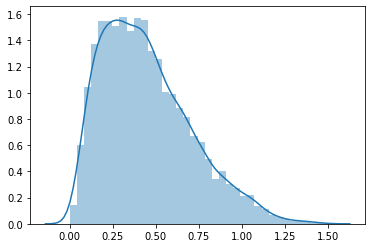}
  \caption{transformed on latent space}\label{fig:eu_limeAfter}
\endminipage\hfill
\minipage{0.43\textwidth}
  \includegraphics[width=\linewidth]{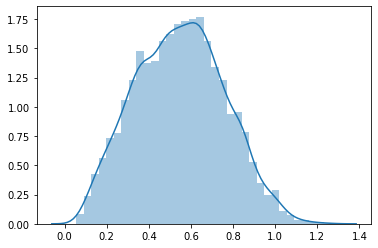}
  \caption{transformed on original space}\label{fig:eu_encToOr)}
\endminipage
\hspace{0.06\textwidth}
\minipage{0.43\textwidth}%
  \includegraphics[width=\linewidth]{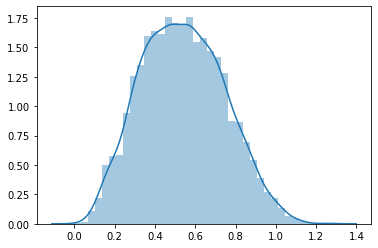}
  \caption{generated on latent space}\label{fig:eu_limeOn}
\endminipage
\end{figure}

Figure~\ref{fig:eu_lime} shows the distribution of the Euclidean distances between $x_i$ and the neighbours generated by LIME in the input space, while Figure~\ref{fig:eu_limeAfter} shows the same distribution of distances between $x^{enc}_i$ and the encoded representations of its neighbours. We observe that the distribution of the neighbourhood distances changes dramatically in the encoded space. In particular, we notice that the distribution of the Euclidean distances between the generated neighbours and the original instance in the encoded space lost its normality, with most of the distances lying between $[.2, .5]$.



By creating the neighbours directly on the latent space, Figure~\ref{fig:eu_limeOn}, we do not jeopardise the latent space adjacency of a neighbour, but we end up with neighbours-instances with abstract, human incomprehensible representations. We proposed one way to deal with this problem, which is to generate the neighbours on the latent space and then decode them to the original space.

\begin{figure}[ht]
\centering
\includegraphics[width=0.57\textwidth]{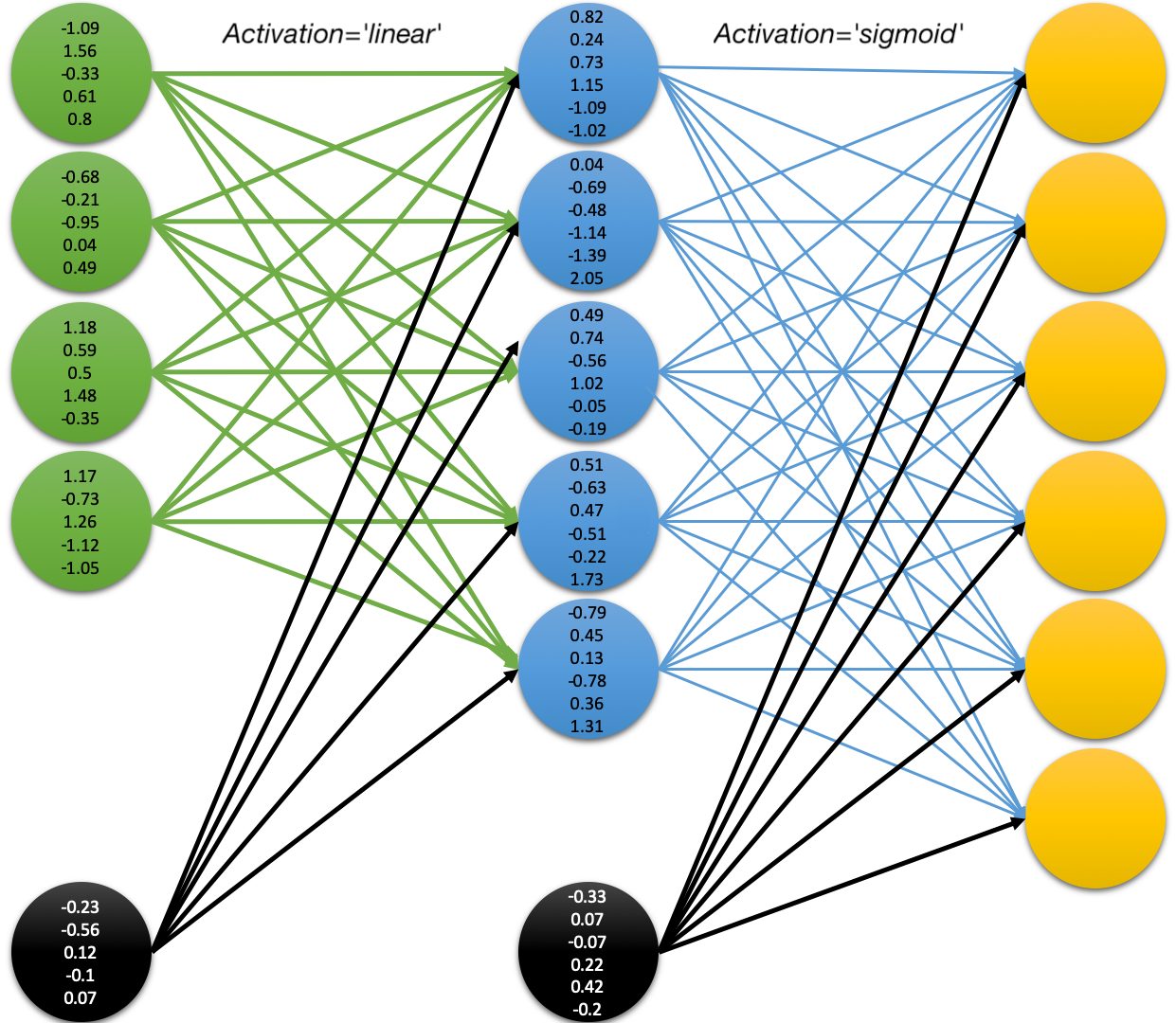}  
\caption{A proposed decoder for the example network in Figure~\ref{simpleNet}} \label{decodersimpleNet}
\end{figure}

Using our simple example in Figure~\ref{simpleNet} as the predictor and encoder, we train a decoder in Figure~\ref{decodersimpleNet} using mean absolute error as loss function. Then, the instance $x_i = [.28, .35, .41, .19, .66, .94]$ is encoded to the abstract representation $x^{enc}_i = [-.2, .25, -.03, .21]$, and finally decoded back to $x_i' = [.28, .39, .43, .19, .68, .91]$. It is visible that the decoder did an impressive job. By building such a successfully trained decoder, we are able to compare Figure~\ref{fig:eu_encToOr)} with Figure~\ref{fig:eu_lime}. It is clear that the distributions in the original space are slightly better, when the neighbour generation process is applied on that space (Figure~\ref{fig:eu_limeOn}). However, when it comes to the abstract space, this is not the case.

\subsection{Transparent Surrogate Model}
The whole process of creating a local neighbourhood around an instance through the NG component and transforming it in the original space using the ND component has the ultimate goal of training an inherently interpretable machine learning model on this neighbourhood in order to provide an explanation for the prediction concerning that instance. Thus, the final component of LioNets is the transparent surrogate model (TS).

\begin{equation}
\label{eq:optimazation}
\begin{aligned}
    & \underset{{\mathcal {L}}_{mae}}{\text{minimise}}
    & & {\mathcal {L}}_{mae}(\mathbf {f(x)} ,\mathbf {g(x)} )\\
    & \text{subject to}
    & & x \in N^{decoded}, \\
    & \text{     }
    & & N^{decoded} \subseteq decoded(N), \\
    & \text{     }
    & & g \in G
\end{aligned}
\end{equation}

In LioNets, we use linear models $G$, specifically Ridge Regression, while we attempt to fit these models to the generated neighbourhood $N^{enc}$ and the prediction probabilities of them, provided by the model being interpreted (Eq.~\ref{eq:optimazation}). There is a tendency to try models such as Lasso or Linear Regression, whose regularisation results in as many features as possible being assigned zero weights. This would result in a smaller set of features that will be more intelligible than larger sets. However, this can lead to worse performance of the classifier in the learning of neighbourhood instances, which is not a desirable attribute, as well as worse interpretations.

\begin{figure}[ht]
\centering
\includegraphics[width=0.9\textwidth]{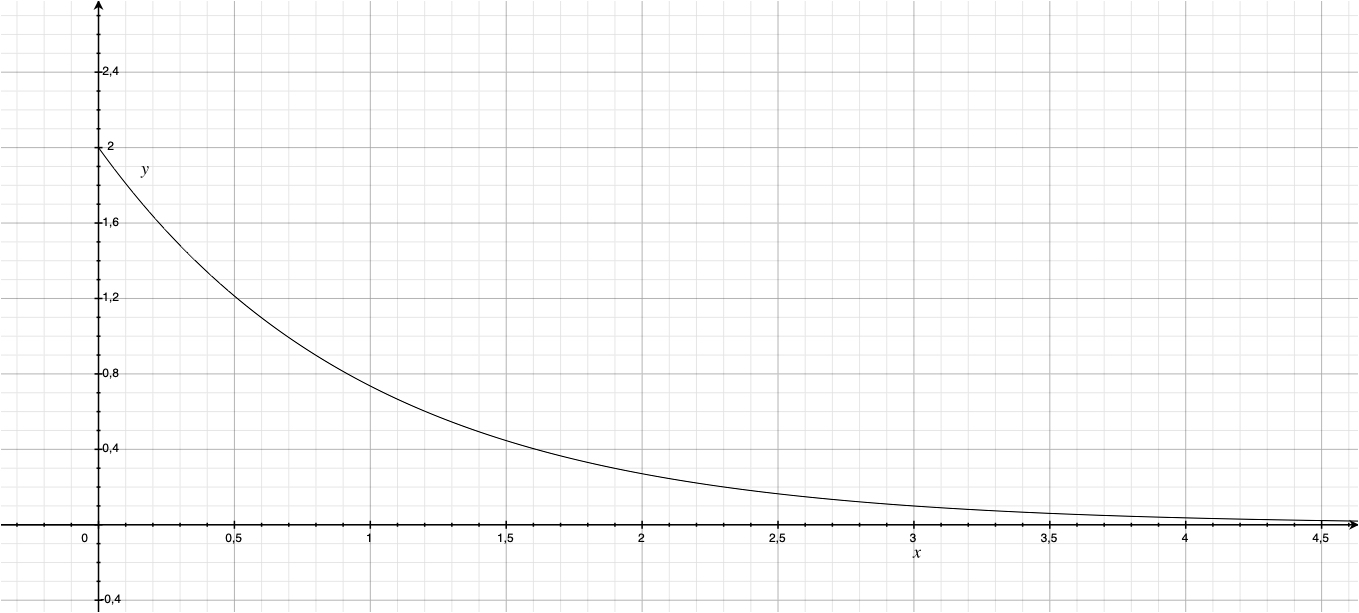}  
\caption{Normalising function for the distances between neighbours and instance. $x= distance$, $y=distance_{normalised}$} \label{kernel}
\end{figure}

Depending on the time available for providing the interpretation, LioNets will adjust from applying a broad grid search for the most suitable parameters for applying a smaller grid search in order to provide faster the interpretation.
In addition, as seen in LIME, a weight is given to each neighbour for TS training, which is the measured Euclidean / Cosine similarity in the encoded space of the neighbour and the original instance, normalised by Eq.~\ref{eq:norm}, where $dimensions$ equal the dimension of abstract space if it is greater than 100 or 100 in any other case. This function favours instances nearest to the original, thus excluding instances with greater distances from the original, as shown in the Figure~\ref{kernel}.

\begin{equation}
\label{eq:norm}
    distance_{normalised} = e^{-distance\frac{log(dimensions)}{2}}log(dimensions)
\end{equation}{} 

\subsection{Explanation Extraction}

Having acquired all the necessary components of LioNets the ultimate goal is to extract from the trained TS model each feature's coefficient. These coefficients can be interpreted as features' importance. Then, it is up to the model designer how to display this kind of information. The default way LioNets presents an explanation is by creating a bar plot, which shows categorical variables (features) and their importances. In Section~\ref{sec:implementation}, we present different ways of visualising this acquired knowledge about the prediction of an instance for textual and time-series data.

In the case of textual data, thanks to the ability to create neighbours in abstract space, TS models will encounter instances (neighbours) that are likely to have features (words) that do not appear in the targeted instance (in the sentence). Thus, the interpretable model will also assign weights to these features, which can be presented as counterfactual words, with respect of their absolute importance (presenting the most positively and negatively important counterfactuals), i.e. words that could potentially have a positive or negative impact on the probability of prediction of the original instance. Finally, because of their relatively small latent distance from the words of the original sentence, these new words will also have a semantic relationship. In Section~\ref{sec:implementation}, we're presenting a few examples of counterfactual words, and how they affect the prediction of a sentence when they are added to it.

\section{Experiments and Evaluation}
\label{sec:implementation}
In order to show the capabilities of LioNets, we have carried out four separate test cases. The first two cases address the issue of binary classification of textual datasets. In the third and fourth scenarios, a time-series dataset is used to solve the issue of binary classification and regression. In this section, we will also propose an extension of faithfulness, and a relaxation of robustness.

The trick of creating neighbours in the latent space enables us to generate neighbourhoods regardless of the nature of the input data. Thus, unlike other methods, there is no restriction on the type of input, which renders LioNets a general method for generating explanations for models trained with simple vector inputs, or even 2D and 3D matrices. The following test cases illustrate this capability, interpreting neural network architectures with simple vectors (first test case), embedding representations (second case) and 2D time-window matrices (third and fourth case) as inputs.

\subsection{Evaluation Setup}
For each test case, we are going to present the implementation of LioNets, a quantitative evaluation according to Table~\ref{tab:experiments}, as well as a qualitative evaluation. In order to provide a quantitative comparison between LioNets and other interpretation techniques, we took into consideration the metrics of Altruist, a relaxed version of robustness, average number of non-zero weights, and fidelity, using mean absolute error (mae) and r-squared ($R^2$). We use regression metrics like MAE and $R^2$, because we measure the fidelity in the probability level, and not in binary scope. The interpretation techniques we tested against LioNets were LIME, Gradient$\times$Input and LRP-e.

\begin{table}[ht]
\centering
\resizebox{\textwidth}{!}{%
\begin{tabular}{|c|c|c|c|c|c|c|c|c|c|} 
\cline{2-9}
\multicolumn{1}{c|}{} & Train & Val & Train & Val & Train & Val & Train & Val & \multicolumn{1}{c}{} \\ 
\cline{2-9}
\multicolumn{1}{c|}{} & \multicolumn{2}{c|}{Altruist} & \multicolumn{2}{c|}{Robustness} & \multicolumn{2}{c|}{NonZero} & \multicolumn{2}{c|}{Fidelity (mae)} & \multicolumn{1}{c}{} \\ 
\hline
LIME & 23.50\% & 21.57\% & 1.03E-03 & 9.57E-04 & \textbf{7.63} & \textbf{7.635} & 1.05E-02 & 1.30E-02 & \multirow{4}{*}{SMS Spam Dataset} \\ 
\cline{1-9}
LioNets & 11.54\% & 19.47\% & 1.22E-03 & 1.29E-03 & 10.66 & 10.71 & \textbf{1.51E-04} & \textbf{6.40E-05} &  \\ 
\cline{1-9}
G$\times$I  & \textbf{3.00\%} & \textbf{3.03\%} & 1.14E-03 & 1.11E-03 & 10.66 & 10.71 & - & - &  \\ 
\cline{1-9}
LRP-e & 48.36\% & 46.08\% & \textbf{6.63E-04} & \textbf{6.60E-04} & 10.66 & 10.71 & - & - &  \\ 
\hline
\multicolumn{1}{c}{} & \multicolumn{1}{c}{} & \multicolumn{1}{c}{} & \multicolumn{1}{c}{} & \multicolumn{1}{c}{} & \multicolumn{1}{c}{} & \multicolumn{1}{c}{} & \multicolumn{1}{c}{} & \multicolumn{1}{c}{} &  \\ 
\hline
LIME & \textbf{43.86\%} & \textbf{46,32\%} & 1.77E-02 & \textbf{1.78E-02} & \textbf{12.60} & \textbf{11.92} & 2.39E-02 & 2.60E-02 & \multirow{4}{*}{Hate Speech Dataset} \\ 
\cline{1-9}
LioNets & 47.85\% & 48,49\% & 3.66E-02 & 3.38E-02 & 17.43 & 15.61 & \textbf{1.96E-03} & \textbf{2.06E-03} &  \\ 
\cline{1-9}
G$\times$I  & - & - & - & - & - & - & - & - &  \\ 
\cline{1-9}
LRP-e & 44.69\% & 47,47\% & \textbf{1.73E-02} & \textbf{1.78E-02} & 17.43 & 15.61 & - & - &  \\ 
\hline
\multicolumn{1}{c}{} & \multicolumn{1}{c}{} & \multicolumn{1}{c}{} & \multicolumn{1}{c}{} & \multicolumn{1}{c}{} & \multicolumn{1}{c}{} & \multicolumn{1}{c}{} & \multicolumn{1}{c}{} & \multicolumn{1}{c}{} &  \\ 
\hline
LIME & 34.04\% & 36.00\% & 7.89E-02 & 7.95E-02 & 700.00 & 700.00 & 8.64E-02 & 6.51E-02 & \multirow{4}{*}{TEDS Binary Classifier} \\ 
\cline{1-9}
LioNets & 33.04\% & 34.57\% & 2.47E-02 & 2.31E-02 & 700.00 & 700.00 & \textbf{3.41E-03} & \textbf{1.38E-03} &  \\ 
\cline{1-9}
G$\times$I  & \textbf{19.75\%} & \textbf{26.00\%} & \textbf{4.33E-03} & \textbf{4.15E-03} & 700.00 & 700.00 & - & - &  \\ 
\cline{1-9}
LRP-e & 36.82\% & 36.36\% & 6.87E-03 & 7.43E-03 & 700.00 & 700.00 & - & - &  \\ 
\hline
\multicolumn{1}{c}{} & \multicolumn{1}{c}{} & \multicolumn{1}{c}{} & \multicolumn{1}{c}{} & \multicolumn{1}{c}{} & \multicolumn{1}{c}{} & \multicolumn{1}{c}{} & \multicolumn{1}{c}{} & \multicolumn{1}{c}{} &  \\ 
\hline
LIME & 38.89\% & 41.21\% & 2.53E-02 & 2.27E-02 & 700.00 & 700.00 & 4.11E-02 & 4.88E-02 & \multirow{4}{*}{TEDS RUL} \\ 
\cline{1-9}
LioNets & 45.79\% & 47.21\% & 2.03E-02 & 1.77E-02 & 700.00 & 700.00 & \textbf{1.26E-02} & \textbf{1.62E-02} &  \\ 
\cline{1-9}
G$\times$I  & \textbf{24.93\%} & \textbf{24.25\%} & 1.44E-02 & 1.43E-02 & 700.00 & 700.00 & - & - &  \\ 
\cline{1-9}
LRP-e & 41.29\% & 37.64\% & \textbf{1.43E-02} & \textbf{1.12E-02} & 700.00 & 700.00 & - & - &  \\
\hline
\end{tabular}}
\caption{The findings of the experiments carried out in the 4 test cases, using 4 different interpretation methods and 4 metrics (The missing values in the fidelity metric for the LRP and Gradient$\times$Input (G$\times$I) is due to the nature of those interpretation techniques, as they do not need to train surrogate models, using only backpropagation)}
\label{tab:experiments}
\end{table}

Altruist was introduced as a metric for evaluating the performance of feature importance interpretation techniques used in machine learning models trained in tabular data. Thus, we extend its applicability to textual and time-series data. In textual data, Altruist will only evaluate the importance assigned to each word of a sentence instead of each word in the vocabulary. The average importance of each sensor will be investigated in time-series data rather than each measurement.

The robustness metric in Eq.~\ref{eq:robustness} demands the identification and use of concepts alongside the explanation provided in a feature importance manner. Moreover, it relies on optimisation and it is computationally heavy, specifically for explanation techniques like LIME. Thus, for our experiments, we use a relaxed version of robustness presented in Eq.~\ref{eq:robustnessRelaxed}, where $L$ is the total number of instances to be examined and $e$ the explanation of an instance $x$. Then, modifying slightly an instance, by subtracting (when in tabular/time-series data) or zeroing (on textual data) the value of the feature with the lowest absolute importance, based on the explanation, we compare the original $e$, which is a vector of $|F|$ values, with the explanation for the prediction of the ``tweaked $x$''. This way robustness will capture how unstable an explanation technique is when a small alteration in the value of the least significant feature happens. 

\begin{equation}
    \label{eq:robustnessRelaxed}
    robustness_{relaxed}= \frac{1}{L}\sum_{i=1}^{L}|e(x_{i}^{original}) - e(x_{i}^{tweaked})|  \text{\\}
\end{equation}

The code of the experiments, the trained models and the material used, are accessible at the GitHub repository ``LionLearn''\footnote{\url{https://git.io/JLmgL}} and the Docker repository\footnote{\url{https://dockr.ly/3qXSoji}}. The results show that LioNets can lead to more precise and truthful explanations than other techniques, on a variety of data types. 

\subsection{Textual Corpora}
The textual data collections we have selected deal with spam SMS detection~\cite{Almeida2011ContributionsResults} and hate speech detection in comments~\cite{mollas2020ethos}, which includes 747 spam and 4,827 ham (non-spam) messages, and 563 comments without and 433 with hate speech content, respectively. For each SMS, the pre-processing consists of the following steps: a) lowercasing, b) stemming and lemmatisation through WordNet lemmatizer \cite{miller1995wordnet} and Snowball stemmer \cite{snowball}, c) phrases transformations (Table~\ref{tab:my-tableTransform}), d) removal of punctuation marks and e) once again, stemming and lemmatisation. For each comment of the hate speech dataset, the preprocessing consists of the following steps: a) lowercasing, b) phrase transformations (Table~\ref{tab:my-tableTransform}) and d) removal of punctuation marks.

\begin{table}[ht]
\centering
\resizebox{0.75\textwidth}{!}{%
\begin{tabular}{|ccc|ccc|ccc|}
\hline
\multicolumn{9}{|c|}{Phrases and words transformations}                          \\ \hline
“what's”  & to & “what is”  & “'ll”    & to & “ will”  & “'s”    & to & “ is”    \\
“don't”   & to & “do not”   & “i'm”    & to & “i am”    & “'ve”   & to & “ have”  \\
“doesn't” & to & “does not” & “he's”   & to & “he is”  & “isn't” & to & “is not” \\
“that's”  & to & “that is”  & “she's”  & to & “she is” & “'re”   & to & “ are”   \\
“aren't”  & to & “are not”  & “it's”   & to & “it is”  & “'d”    & to & “ would” \\
“\%”      & to & “ percent” & “e-mail” & to & “e mail” &         &    &          \\ \hline
\end{tabular}}
\caption{Phrases and words transformations}
\label{tab:my-tableTransform}
\end{table}

\subsubsection{SMS Spam Dataset with TFIDF Representations}
The first case concerns the issue of SMS Spam detection. To address this problem of binary classification, to identify SMSs without (ham) or with spam content, we will need to transform the sentences into vectors. We will use the vectorization technique of Term Frequency-Inverse Document Frequency (TFIDF)~\cite{sparck1972statistical}, which is widely acknowledged by many researchers because it is simple and efficient. 

\begin{figure}[!htb]
\centering
\caption*{Architectures of networks for SMS Spam dataset:}
\minipage{0.35\textwidth}
  \centering
  \includegraphics[width=0.5\linewidth]{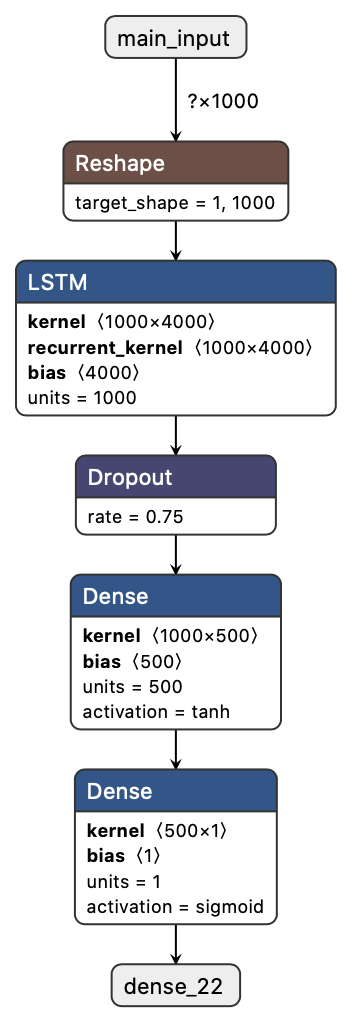}
  \caption{The predictor}\label{fig:FirstCasePredictor}
\endminipage \hspace{0.1in}
\minipage{0.35\textwidth}
  \centering
  \includegraphics[width=0.49\linewidth]{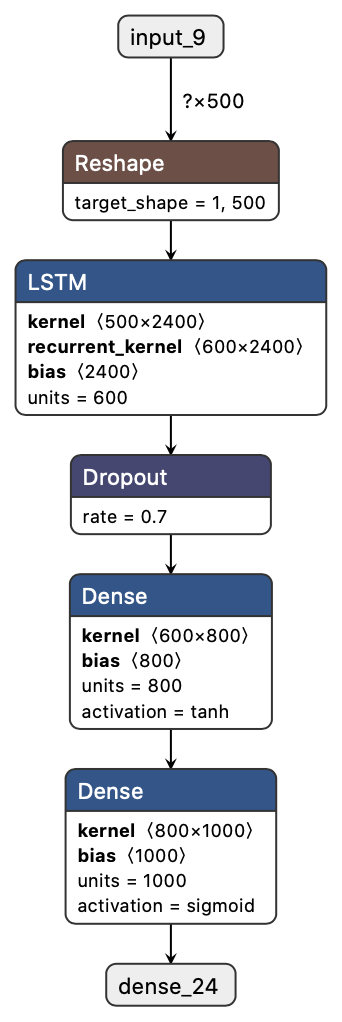}
  \caption{The decoder}\label{fig:FirstCaseDecoder}
\endminipage\hfill
\end{figure}

The predictor (Figure~\ref{fig:FirstCasePredictor}) learned to distinguish the types of messages (spam or ham) using 80\% of the dataset as training data and 20\% as test data. In predictor's training, the optimiser ``adam'' and the loss function ``binary crossentropy'' were utilised. The model's performance in terms of $F_1$-score (`macro-averaged') was 99.90\% on the training set and 96.62\% on the test set, while the results were 99.83\% and 95.43\% respectively in terms of balanced accuracy. 

The decoder (Figure~\ref{fig:FirstCaseDecoder}) uses as input the encoded representation of instances extracted from the penultimate layer of the predictor, with 500 dimensions, and it has output the original shape of the instances, with 1000 dimensions. Using ``adam'' as optimiser and ``binary crossentropy'' as loss function, we acquired 0.0074 and 0.0098 error, as well as 0.0020 and 0.0024 `mean absolute error', on the training set and test set, respectively. Since the dataset was relatively small (5574 instances), we've used unsupervised data to enhance the performance of the decoder.

Here are some examples of the decoder:
\begin{lstlisting}[caption=Examples of decoded sentences of SMS Spam dataset]
 Train Data:
"""""""""""""""""""""""""""""""""""""""""""""""""""""""""""""""""""""""""""""""""
   Original: also am but do have in onli pay to
    Decoded: also am but do have in not onli pay to
"""""""""""""""""""""""""""""""""""""""""""""""""""""""""""""""""""""""""""""""""
   Original: come got it me now ok or then thk wan wat
    Decoded: come got it me now ok or then thk wan wat
"""""""""""""""""""""""""""""""""""""""""""""""""""""""""""""""""""""""""""""""""
 Test Data:
"""""""""""""""""""""""""""""""""""""""""""""""""""""""""""""""""""""""""""""""""
   Original: been better day each even give god great more never reason thank to
    Decoded: been better day for give god great it more thank the to
"""""""""""""""""""""""""""""""""""""""""""""""""""""""""""""""""""""""""""""""""
   Original: am at be late there will
    Decoded: am at be late there will
"""""""""""""""""""""""""""""""""""""""""""""""""""""""""""""""""""""""""""""""""
\end{lstlisting}

Comparing LioNets with other methodologies in a quantitative manner, we can observe in Table~\ref{tab:experiments} significant superiority of the Gradient$\times$Input algorithm in terms of the Altruist score. This is justified by the fairly simple architecture of the predictor. Nevertheless, LRP cannot operate correctly due to the presence of recurrent networks, even if the network uses activation functions that comply with the LRP requirements (hyperbolic tangent---tanh). Moreover, LioNets perform better than LIME and LRP in terms of this score. As far as other metrics are concerned, LRP achieves the highest robustness scores, LIME provides the smallest explanations, and LioNets produces linear models that better approximate neural network predictions than LIME.

\begin{figure}[ht]
\centering
\caption*{SMS: ``congrat treat pend am not on mail for day wil mail onc thru respect mother at home check mail''}
\minipage{0.43\textwidth}
  \includegraphics[width=\linewidth]{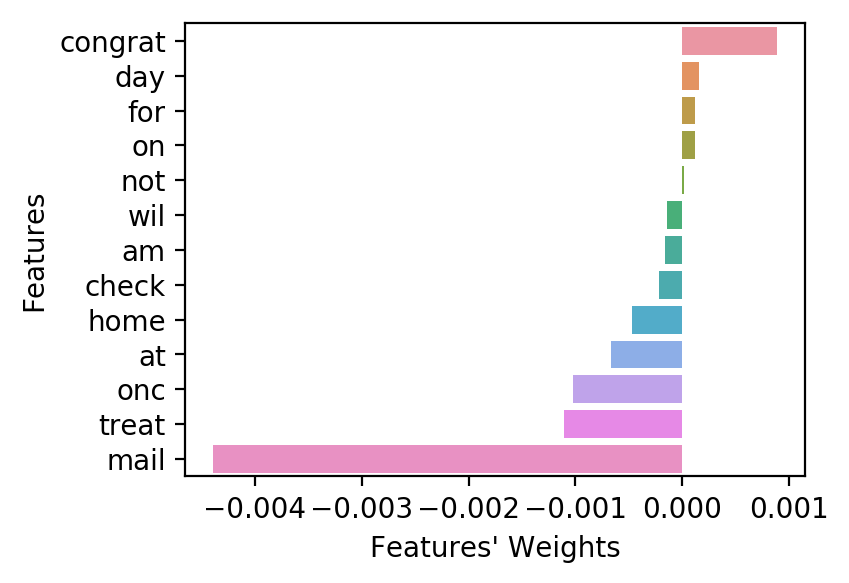}
  \caption{Features appearing in the instance}\label{fig:FirstCaseExampleExplanation}
\endminipage \hspace{0.3in}
\minipage{0.425\textwidth}
  \includegraphics[width=\linewidth]{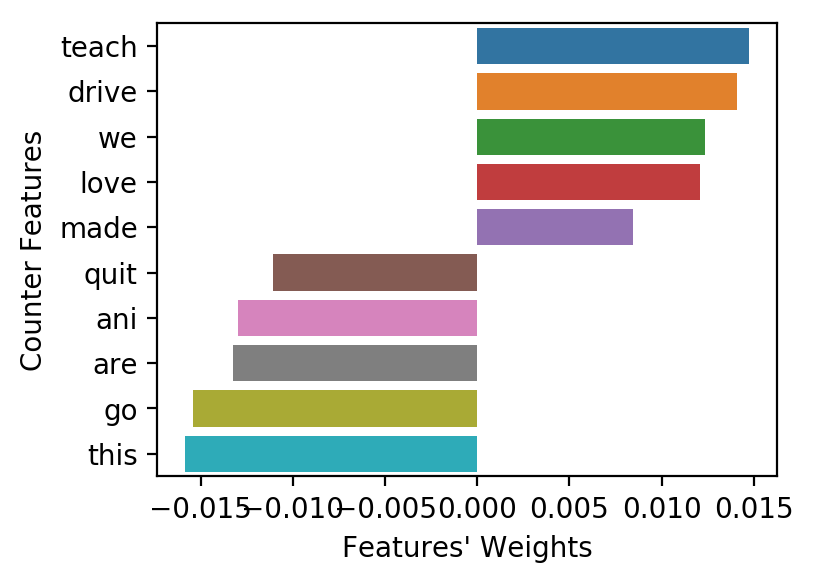}
  \caption{Features not appearing in the instance}
  \label{fig:FirstCaseExampleCounterExplanation}
\endminipage\hfill
\end{figure}

By training all the necessary components of LioNets, it is feasible to apply the technique to an instance's classification to extract interpretations to qualitatively evaluate the technique. To give an illustration, for the instance: \emph{``congrat treat pend am not on mail for day wil mail onc thru respect mother at home check mail''} with probability to contain spam $10.27\%$, we can assign importance to each of the features of the instance (Figure~\ref{fig:FirstCaseExampleExplanation}), as well as we can identify features not appearing in the instance but found in the neighbourhood that may influence the prediction (Figure~\ref{fig:FirstCaseExampleCounterExplanation}).

In addition, in order to determine the quality of the explanations we removed the word ``congrat'', and we observed a decreased in the probability to 10.15\%, while by removing the word ``mail'', we observed an increase in the likelihood to 10.36\%. Another noteworthy feature of LioNets is the ability to explore features appearing in an instance's neighbourhood and present them as counter features (Figure~\ref{fig:FirstCaseExampleCounterExplanation}), when handling sparse data, such as text data. 

We extracted the local neighbourhood for the instance, and by observing the counter features, which created through the process of neighbourhood generation in the latent space, we added to the sentence the word \emph{``teach''} and the probability raised (10.29\%), while when we added the word \emph{``this''} the probability decreased (10.23\%). This uncovers an interesting fact, that in this local prediction the word \emph{``teach''} is contributing to the ``spam'' class, while in a sentence like the following: \emph{``\textbf{teach} me app da when you come to colleg''}, the word ``teach'' is the most influencing feature to the ``ham'' class.

\begin{figure}[ht]
\centering
\caption*{Architectures of networks for Hate Speech dataset:}
\minipage{0.64\textwidth}
  \centering
  \includegraphics[width=\linewidth]{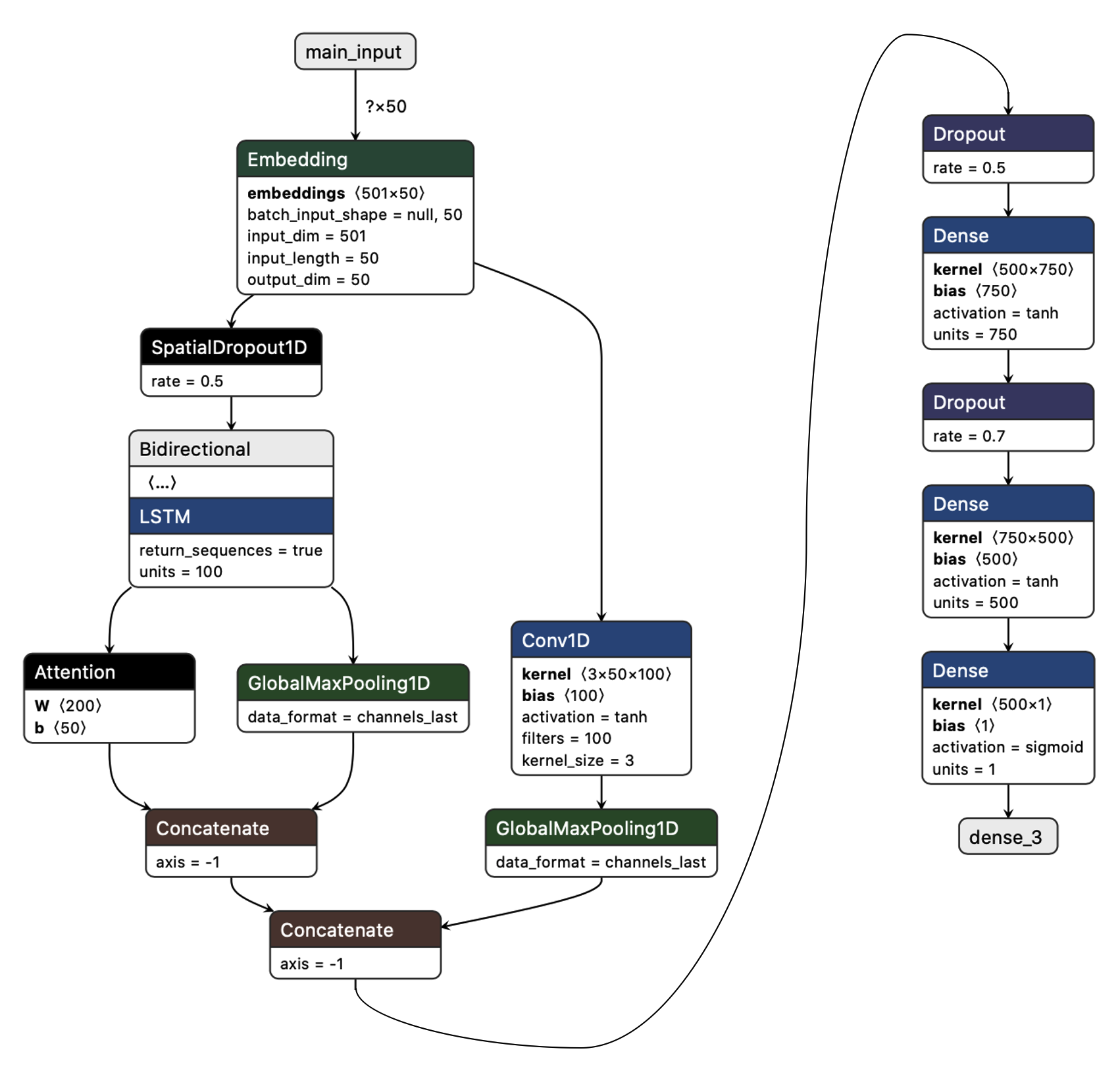}
  \caption{The predictor}\label{fig:SeconfCasePredictor}
\endminipage \hspace{0.1in}
\minipage{0.275\textwidth}
  \centering
  \includegraphics[width=0.75\linewidth]{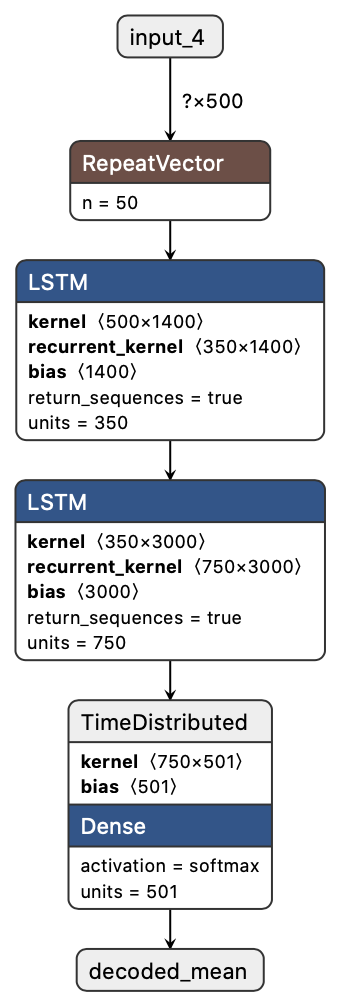}
  \caption{The decoder}\label{fig:SeconfCaseDecoder}
\endminipage\hfill
\end{figure}

\subsubsection{Hate Speech dataset with Embedding Representations}
The second case is about hate speech detection. The ``ETHOS'' dataset~\cite{mollas2020ethos} contains comments on social media platforms that may or may not contain hate speech. In order to train a model to predict the appearance of hate speech content in a sentence, we have transformed the input sentences into embedding matrices. For each sentence, we kept only 50 words (max\_words), and by using pre-trained GloVe embeddings~\cite{glove}, we transformed each word into a 50-dimensional vector resulting in an embedding matrix of 50$\times$50. We have set the vocabulary to a limit of 500.

Hate speech detection is a very challenging task, making the architecture of the neural network used more complicated. The predictor (Figure~\ref{fig:SeconfCasePredictor}) was trained using 80\% of the data as a training set and 20\% as a test set. Using the ``binary crossentropy'' as loss function and the ``adam'' optimiser, it reached a performance of 77.86\% and 70.71\% in terms of $F_1$-score (`macro-averaged'), and 77.25\% and 70.46\% in terms of balanced accuracy, for the train and test set, respectively.

The decoder (Figure~\ref{fig:SeconfCaseDecoder}) using as input the encoded representation of instances extracted from the penultimate layer of the predictor, with 500 dimensions, it outputs a 50$\times$501 matrix, where 50 are the number of words of the sentence, and 501 the words of the vocabulary (the additional one to 500 words is the `UKN'). Using ``adam'' as optimiser and ``categorical crossentropy'' as loss function, we acquired 0.1472 and 0.1152 error, on the training set and test set, respectively. Since the dataset was very small (998 instances), we've used both train and test data to train the decoder.

Below, we present some examples of the decoded sentences:
\begin{lstlisting}[caption=Examples of decoded sentences of Hate Speech dataset]
 Train Data:
""""""""""""""""""""""""""""""""""""""""""""""""""""""""""""""""""""""""""""
    Original: maybe it is because you are autistic wish you death 
     Decoded: maybe it is because you are autistic wish you death 
""""""""""""""""""""""""""""""""""""""""""""""""""""""""""""""""""""""""""""
    Original: disgusting sick UKN kill yourself 
     Decoded: disgusting sick UKN kill yourself 
""""""""""""""""""""""""""""""""""""""""""""""""""""""""""""""""""""""""""""
 Test Data:
""""""""""""""""""""""""""""""""""""""""""""""""""""""""""""""""""""""""""""
    Original: UKN is the UKN that UKN out of UKN UKN UKN 
     Decoded: UKN is the UKN that UKN out of UKN UKN 
""""""""""""""""""""""""""""""""""""""""""""""""""""""""""""""""""""""""""""
    Original: what the fuck stupid people 
     Decoded: what the fuck stupid people 
""""""""""""""""""""""""""""""""""""""""""""""""""""""""""""""""""""""""""""
\end{lstlisting}

Gradient$\times$Input does not take place in this test case due to the limitations of its implementation. LIME was the winner of the best Altruist score, while LioNets and LRP-e were pretty close. LRP-e provides the most stable interpretations, in terms of robustness, with LIME to follow. LioNets had the worst performance in the robustness metric. In addition, we observe that LioNets has achieved a higher fidelity score than LIME. Finally, only LIME was able to reduce the number of non-zero weights by 4.26 features on average.

\begin{figure}[!htb]
\centering
\caption*{Comment: ``or maybe just do not follow UKN UKN religions from the UKN UKN''}
\minipage{0.43\textwidth}
  \centering
  \includegraphics[width=\linewidth]{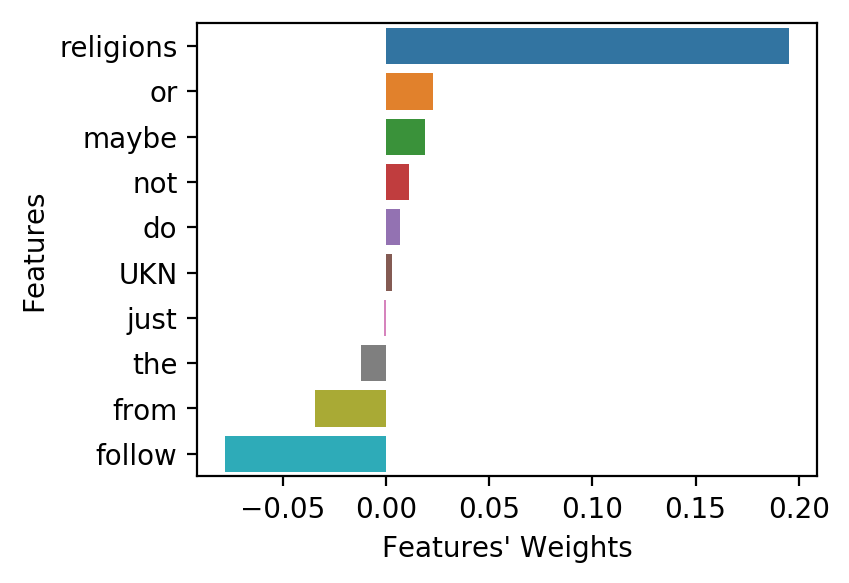}
  \caption{Features appearing in the instance}\label{fig:SecondCaseExampleExplanation}
\endminipage \hspace{0.3in}
\minipage{0.44\textwidth}
  \includegraphics[width=\linewidth]{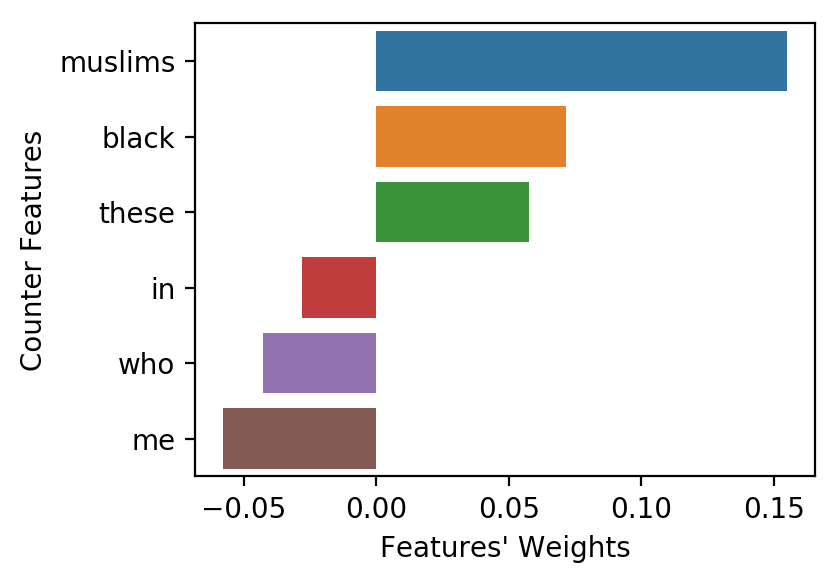}
  \caption{Features not appearing in the instance}
  \label{fig:SecondCaseExampleCounterExplanation}
\endminipage\hfill
\end{figure}

In order to examine the qualitative performance of LioNets in this dataset, we present an explanation provided by the technique, for the following comment: \emph{``or maybe just do not follow UKN UKN religions from the UKN UKN''}. The neural network predictor originally assigns a $46.48\%$ probability to contain hate speech content, which means is uncertain. The explanation is shown in Figure~\ref{fig:SecondCaseExampleExplanation} and the counterexamples identified are shown in Figure~\ref{fig:SecondCaseExampleCounterExplanation}.

This particular example is correctly unidentified by the neural network. Due to the relatively small size of the vocabulary, the word before the word \emph{``religions''} is unknown (UKN) and therefore the predictor cannot judge the comment as hateful. It is obvious that the word \emph {``religions''} increases the probability that the comment contains hate speech content, and by removing it the probability is reduced to $14.603\%$. 

Taking advantage of the alternative ability of LioNets, to assign feature importance to words not appearing in the original sentence, we attempt to add in the sentence the word \emph{``muslims''} in the position of the first UKN word, and we observe a radical increase in the probability from $46.45\%$ to $61.95\%$. Sequentially, we try the same experiment by adding the word \emph{``me''}, which leads to a drop of $22.47\%$ in the probability. Such experiments are very useful to the model designer. He may note that stopwords like `in', `who' and `these' influence output a lot, so it might be prudent to consider excluding them from the training data.

\subsection{Turbofan Engine Degradation Simulation Dataset --- Time-Series}
\label{TEDS}

The next two test cases concern the Turbofan Engine Degradation Simulation dataset~\cite{saxena2008damage,saxena2008turbofan}. This dataset contains four individual simulated datasets. Each one of the four datasets consists of multiple multivariate time series about different engine units and their degree of wear. Particularly, for every unit there are time-steps accompanied by the remaining useful lifetime. We chose the first sub dataset to create both a binary classifier and a remaining-useful-lifetime (RUL) predictor. In this section, we are going to describe the process of applying LioNets and extracting explanations from the models.

\begin{figure}[!htb]
\centering
\caption*{Architectures of networks for Turbofan Engine Degradation Simulation dataset:}
\minipage{0.255\textwidth}
  \includegraphics[width=\linewidth]{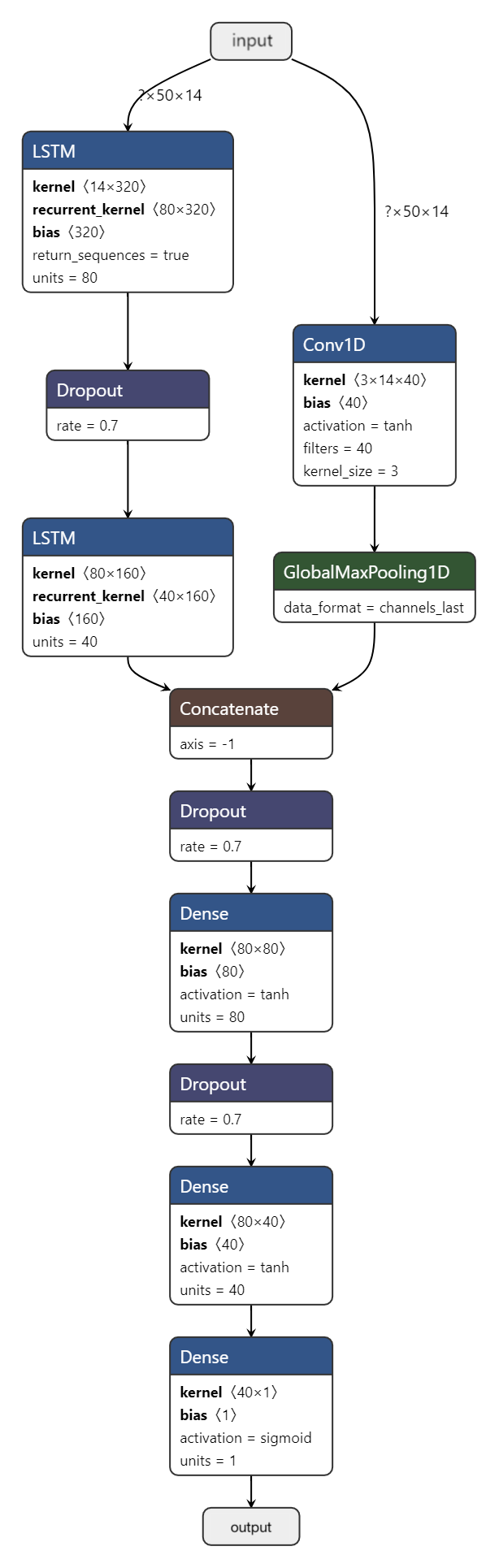}
  \caption{The predictor}\label{fig:FourthCasePredictor}
\endminipage \hspace{1in}
\minipage{0.233\textwidth}
  \includegraphics[width=0.92\linewidth]{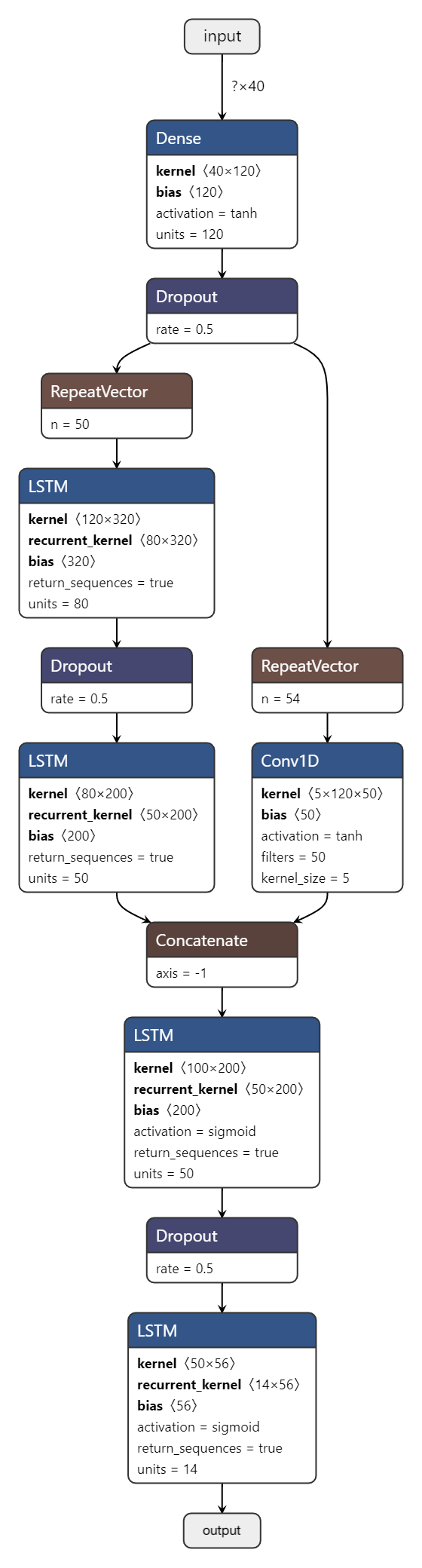}
  \caption{The decoder}\label{fig:FourthCaseDecoder}
\endminipage\hfill
\end{figure}

Before we proceed to the models' design, we have to apply some feature engineering in the input data. We firstly discarded some features, like the third operational setting and some sensors $[1, 5, 6, 10, 16, 18, 19, 22, 23, 24, 25, 26]$, because they did not contain any information (all values were NaN, or they had the exact same value). Then, inspired by a recent work~\cite{RULotherPaper}, we removed the two additional operating settings. In the end, we keep 14 sensor measurements for each time-step of every unit. Lastly, we are setting a time-window parameter $N \in \mathbb{N}$. The neural predictor will be trained to approximate the remaining useful lifetime (RUL) of one unit, having as input the measurements of some sensors of the current time-step, accompanied by the records of the $N-1$ previous time-steps. 

This specific dataset will give us the ability to present the LioNets' effectiveness to explain models trained on complex input shapes. For both the classification and the regression models, we designed and used the same architectures for the predictor (Figure~\ref{fig:FourthCasePredictor}) and the decoder (Figure~\ref{fig:FourthCaseDecoder}).

LioNets uses the predictor, the encoder, which is extracted from the predictor, as well as the decoder to create a neighbourhood for a given instance. The input shape is a 2D matrix of $50\times14$. In order to use the generated neighbourhood to train a transparent linear model, we are reshaping the neighbours from $50\times14$ to $700\times1$, thus, we are transforming the matrices to vectors. Then, by training the linear model and extracting the coefficients we have 700 different values. 

We propose to display this knowledge in several ways. The first way is to display the mean average influence of each sensor over the $N$ time-steps. Lastly, a particular sensor can be selected and the influence of the $N$ time steps as well as the sensor's value can be observed in a plot.

\subsubsection{Explanation on Binary Classification}
This dataset innately concerns a regression problem. Thus, in order to build a binary classifier, we need to transform it into a classification problem. We will use a time-threshold $T$, which is going to transform the RUL of each time-step of all units to a binary value $\in \{0,1\}$. Specifically, we set the value of a time-step to $0$ when the condition $RUL > T$ is true, and to $1$ when the $RUL \leq T$. We trained the predictor, which has the architecture of the encoder of Figure~\ref{fig:FourthCasePredictor} with an output layer with sigmoid function, with the ``adam'' optimiser and the ``binary crossentropy'' loss function. The performance of the model in terms of $F_1$-score (``weighted'') was 0.9473, and balanced accuracy was 0.8559. The decoder (Figure~\ref{fig:FourthCaseDecoder}) was trained with the ``adam'' optimiser and the ``root mean squared error---rmse'' loss function. The performance of the model in terms of the root mean squared error was 0.0683, while the mean absolute error was 0.0521. 

In Table~\ref{tab:experiments}, we can observe that Gradient$\times$Input achieves the highest Altruist score, while LioNets is following. Gradient$\times$Input have also achieved the lowest robustness score, while LioNets performed better than LIME, in terms of fidelity. However, neither technique managed to reduce the number of non-zero weights.

\begin{figure}[!htb]
\centering
\caption*{Explanations of classifier on Turbofan Engine Degradation Simulation dataset:}
\minipage{0.75\textwidth}
  \includegraphics[width=\linewidth]{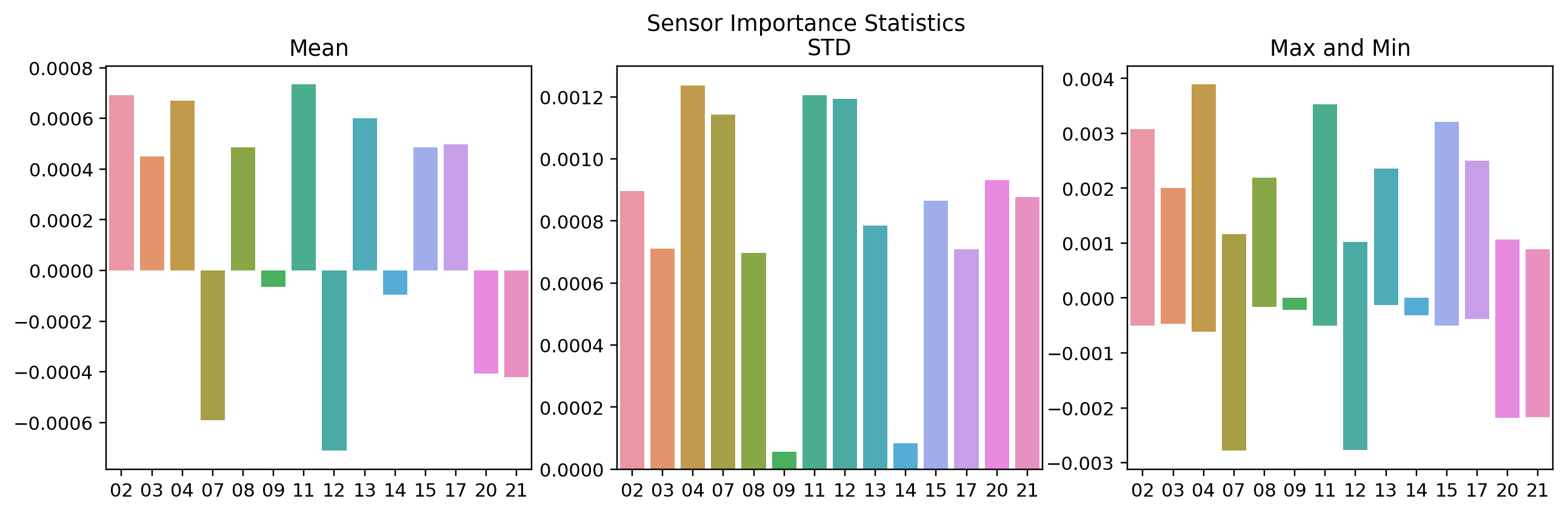}
  \caption{Mean, STD and Max/Min influence per sensor}
  \label{fig:FullTEDS_Class_explanation1}
\endminipage \hfill
\minipage{0.75\textwidth}
  \includegraphics[width=\linewidth]{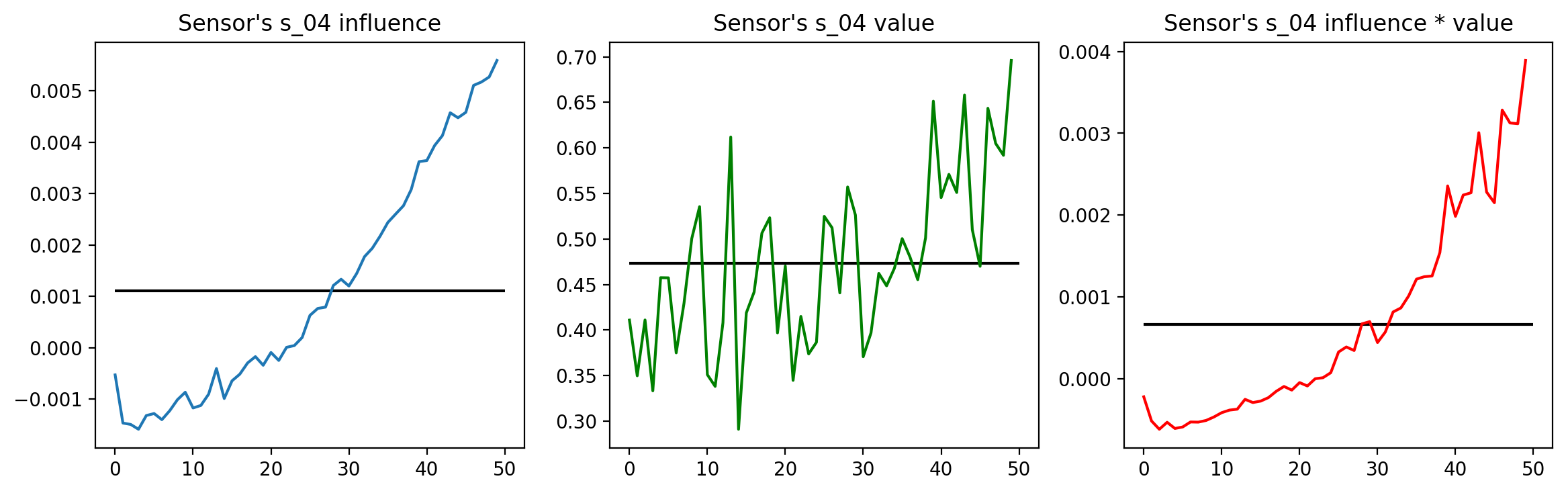}
  \caption{Influence, value and influence$\times$value of each time step of sensor 4}\label{fig:FullTEDS_Class_explanation2}
 \endminipage\hfill
\end{figure}

We may perform a qualitative assessment after all the required components have been assembled, with a properly trained predictor, encoder and decoder. In a random instance, the predictor assigns 56.02 percent of the probability that the component may need to be maintained. We would like to have a lower probability, below 50\%, to conclude that the component does not need maintenance. We use the LioNets technique to obtain a vector of $14 \times 50 = 700$ values. For each time-step of each sensor, we have an influence factor. The first way to display this information is to aggregate the 50 influence values for each sensor and present their mean, STD and max / min value (Figure~\ref{fig:FullTEDS_Class_explanation1}). 

In addition, we select one of the most important sensors from these plots. The 4$^{th}$ sensor tends to have a greater impact on the classifier by positively influencing the ``need maintenance'' class. The last 10 measurements of the 4$^{th}$ sensor are modified (subtracted by 0.1) because, according to Figure~\ref{fig:FullTEDS_Class_explanation2}, these measurements had a high impact. We observe that these values are also higher than the average. After these changes, the neural network assigned a likelihood of 35.34\% to the ``need maintenance'' class. Thus, if we had decreased the measurements of this sensor in these time steps, we might have reduced the likelihood of failure.

\subsubsection{Explanation on Remaining Useful Lifetime Estimator}
In contrast to the binary classifier, the output of the remaining useful lifetime estimator is $Y \in \mathbb{R}_{\ge 0}$. The predictor from Figure~\ref{fig:FourthCasePredictor} was trained with the ``adam'' optimizer and the ``root mean squared error---rmse'' loss function. The performance of the model in terms of the root mean squared error was 32.11, while the mean absolute error was 22.96. The decoder from Figure~\ref{fig:FourthCaseDecoder} was trained with the ``adam'' optimizer and the ``root mean squared error---rmse'' loss function. The performance of the model in terms of the root mean squared error was 0.0767, while the mean absolute error was 0.0595. 

In this last test case, Gradient$\times$Input scores the best Altruist score, while LioNets was the second-best technique. LIME and LRP-e are struggling to provide truthful explanations. LRP-e provides the most stable, in terms of robustness, interpretations, with Gradient$\times$Input and LioNets to follow. In addition, we observe that LioNets has achieved a better fidelity score, outperforming LIME, which has also achieved high fidelity scores. However, neither technique managed to reduce the number of non-zero weights.

\begin{figure}[!htb]
\centering
\caption*{Explanations of RUL predictor on Turbofan Engine Degradation Simulation dataset:}
\minipage{0.75\textwidth}
  \includegraphics[width=\linewidth]{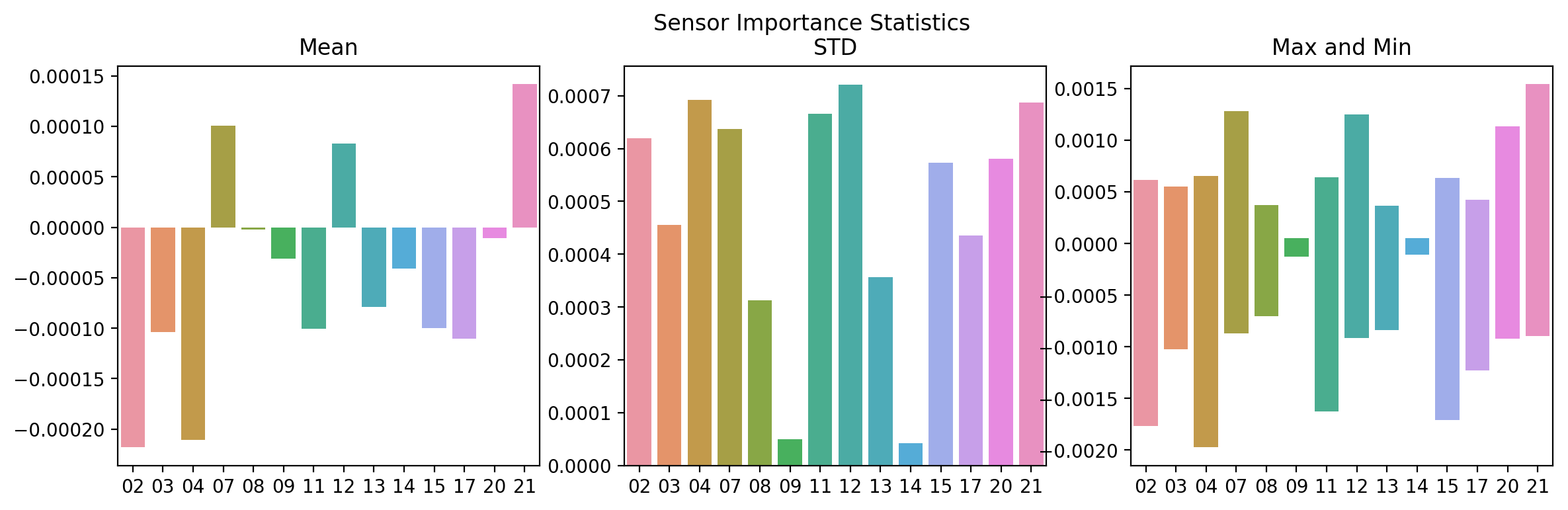}
  \caption{Mean, STD and Max/Min influence per sensor}
  \label{fig:FullTEDS_RUL_explanation1}
\endminipage \hfill
\minipage{0.75\textwidth}
  \includegraphics[width=\linewidth]{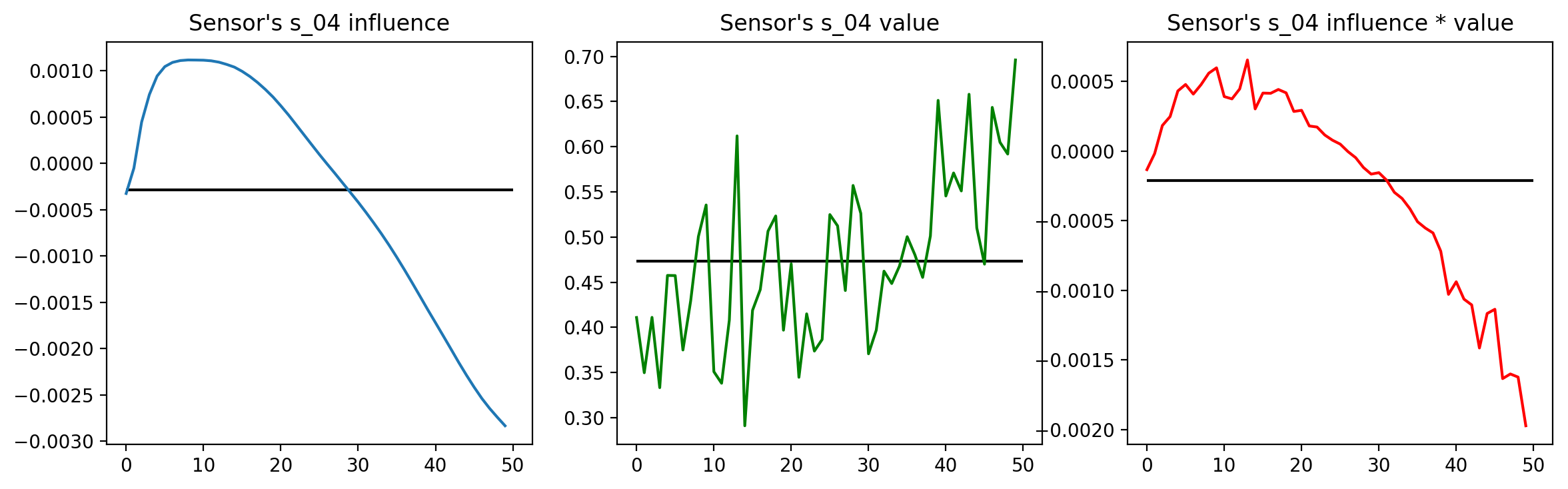}
  \caption{Influence, value and influence$\times$value of each time step of sensor 4}\label{fig:FullTEDS_RUL_explanation2}
 \endminipage\hfill
\end{figure}

Then we choose an instance, a set of measurements for the sensors, which we know will result in a low RUL value. The neural network accurately predicts that the remaining useful life of the component is 25.04. The LioNets interpretation technique is then applied to this prediction, and the local linear model prediction is 25.21, which is very close to the neural network's prediction. Moreover, we generate the feature importance plots in the Figure~\ref{fig:FullTEDS_RUL_explanation1}. From these plots, we can see that the measurements of the 4$^{th}$ sensor influence negatively most the prediction. Thus, the measurements of the sensor are adjusted by decreasing the values from the 35$^{th}$ time-step and afterwards, as if these measurements had the most negative effect on the prediction according to Figure~\ref{fig:FullTEDS_RUL_explanation2}, and at the same time it has higher than normal reported values. Finally, the prediction for the updated measurements from the neural network is rising to 33.57. We have therefore managed to extend the lifetime of the component using this interpretation.

\section{Discussion}
We have reached the following research findings through the experiments which support the initial claim that LioNets is a practical and complete technique to be used in a plethora of applications. As far as the truthfulness (Altruist score) of the interpretations is concerned, LioNets cannot achieve the efficiency of the G$\times$I algorithm, but in 2 out of 4 cases it exceeds the performance of the LRP and LIME algorithms. Both LRP and G$\times$I techniques have the highest robustness scores of all tests, while LioNets follows. LIME seems to have less stable interpretations. Furthermore, LIME is the only methodology that reduces the features appearing in the interpretations, at least in the textual test cases. Finally, between LioNets and LIME, LioNets had the highest fidelity score in any experiment.

Through the qualitative experiments, LioNets showcased as a pretty useful tool to try to change the prediction for an instance. Indeed, in the textual datasets we found the important words, which when they were removed the prediction was changing, as well as semantically similar to the context words, which when were added to the sentence the prediction was changing as well. Moreover, in the time-series test cases, the interpretations were able to lead to conclusions on how a low estimated lifetime prediction could have been prevented.

So why should anyone prefer LioNets over techniques like G$\times$I? In almost every case, G$\times$I exceeds all the other techniques. However, its implementation's limitations are in contrast to the applicability of LioNets. In addition, neighbourhood techniques allow users to experiment and try out different ways to extract interpretations. For example, a set of counterfactual words with semantic similarity to an instance can be extracted through LioNets generated neighbours specifically in textual datasets.

Comparisons are almost equal between LioNets and LIME in terms of truthfulness (Altruist score), with LIME achieving 35.67\% and LioNets 35.99\% on average for both train and test sets in every test case. Providing surrogate models with better fidelity, LioNets also generates richer semantically neighbours that are closer to the encoded representation of the original instance in contrast to LIME's neighbourhoods. In textual datasets, the excellence of LioNets over LIME is even more evident because of the ability to identify words that do not appear in the original sentence.

Therefore, LioNets can be specified as the interpretation medium for a neural network with a slight trade-off of truthfulness and almost no trade-off of robustness, offering richer explanations and counter-words in textual datasets. In addition, LioNets producing a local neighbourhood enables users to work to identify new ways to explain an instance. For example, the relationship and impact of word pairs and words' positions on textual datasets can be exploited through these neighbourhoods. Finally, such neighbourhoods can be used in a time-series dataset to train various forms of surrogate models that export details that are stronger and more useful.

\section{Conclusion}

In summary, we provided a detailed presentation of the LioNets architecture, which offers accurate and consistent interpretations of neural network decisions that are comparable to other state-of-the-art techniques. This technique ensures a better relationship between the developed neighbours of the instance, as the generation process is carried out on the penultimate layer of the network. In this space, neighbours have lower dimensions, and phenomena such as the \textit{curse of dimensionality} are better treated, while at the same time, neighbours have richer semantic information for the model itself. In addition, we have shown the ability of LioNets to adapt to various data types (textual and time series). 

The validity of this research is assessed by quantitative and qualitative experiments using well-known metrics in four separate test cases. We have implemented an extension of two metrics to improve the quantitative evaluation, relaxed robustness and Altruist on textual and time-series data. New methods for visualising the interpretations of LioNets have also been introduced. A new way of finding counterfactual terms in a sentence has also been presented in the textual test cases.

However, one of the main drawbacks of LioNets is that it focuses exclusively on the interpretation of neural networks, so it is not a model agnostic technique. Also, the overall process of using LioNets is more complex than other approaches, since it requires the preparation of a decoder, which is a difficult task most of the time. Future research plans include introducing LioNets for multi-class or multi-label tasks such as Image Recognition or Object Detection. Finally, we intend to build an extension of LioNets that will not require a separate decoder to be provided by the user.

\section*{Acknowledgment}
This paper is supported by the European Union's Horizon 2020 research and innovation programme under grant agreement No 825619 [AI4EU Project]\footnote{\url{https://www.ai4eu.eu}}.

\bibliographystyle{unsrt}

\end{document}